%% file: vln_behave_draft.tex
\PassOptionsToPackage{usenames,dvipsnames}{xcolor}

\documentclass[10pt,twocolumn,letterpaper]{article}

\usepackage{cvpr}      %

\usepackage{graphicx}
\usepackage{amsmath}
\usepackage{amssymb}
\usepackage{booktabs}
\usepackage{colortbl}
\usepackage{xcolor}
\usepackage{dblfloatfix}
\usepackage{paralist}
\usepackage[accsupp]{axessibility}  %
\usepackage[pagebackref,breaklinks,colorlinks]{hyperref}
\usepackage{longtable}
\usepackage{appendix}
\usepackage{subfiles}
\usepackage[capitalize]{cleveref}
\crefname{section}{Sec.}{Secs.}
\Crefname{section}{Section}{Sections}
\Crefname{table}{Table}{Tables}
\crefname{table}{Tab.}{Tabs.}

\PassOptionsToPackage{usenames,dvipsnames}{xcolor}

\def\cvprPaperID{9851} %
\def\confName{CVPR}
\def\confYear{2023}

\graphicspath{{figures/}{../figures/}}
\input{macros.tex}

\begin{document}

\title{Behavioral Analysis of Vision-and-Language Navigation Agents}

\author{Zijiao Yang\\
Oregon State University\\
{\tt\small yangziji@oregonstate.edu}
\and
Arjun Majumdar\\
Georgia Institute of Technology\\
{\tt\small arjun.majumdar@gatech.edu}
\and
Stefan Lee\\
Oregon State University\\
{\tt\small leestef@oregonstate.edu}
}
\maketitle
% add page number
\thispagestyle{plain}
\pagestyle{plain}

\begin{abstract}
\input{sections/0_abstract}

\end{abstract}

\input{sections/1_introduction}

\input{sections/2_related}

\input{sections/4_methodology}

\input{sections/5_experiments}
\input{sections/6_discussion}

{\small
\bibliographystyle{ieee_fullname}
% \bibliography{egbib}

\input{bib}
}
\newpage
\section*{Supplementary material}
\begin{appendix}
\input{supplementary/discussion}

\input{supplementary/stats}

\input{supplementary/results_extended}
\input{supplementary/templates}
\input{supplementary/teacher_forcing_effects}

\end{appendix}
\end{document}

%% file: macros.tex
\newcolumntype{s}{>{\columncolor[gray]{.85}[.5\tabcolsep]}c}

\definecolor{pltgreen}{RGB}{0,128,0}  %

\newcommand{\supp}[1]{\textcolor{PineGreen}{[SUPP] #1}}
\renewcommand{\supp}[1]{#1}

\newcommand{\tightcaption}{\vspace{-0.5\baselineskip}}
\newcommand{\xhdr}[1]{\vspace{2pt}\noindent\textbf{#1}}
\newcommand{\xhdrflat}[1]{\noindent\textbf{#1}}

\newcommand{\tabref}[1]{Tab.~\ref{#1}\xspace}
\newcommand{\figref}[1]{Fig.~\ref{#1}\xspace}
\newcommand{\secref}[1]{Sec.~\ref{#1}\xspace}

\newcommand{\myquote}[1]{\emph{``#1''}}

%% file: sections/0_abstract.tex
To be successful, Vision-and-Language Navigation (VLN) agents must be able to ground instructions to actions based on their surroundings. In this work, we develop a methodology to study agent behavior on a skill-specific basis -- examining how well existing agents ground instructions about stopping, turning, and moving towards specified objects or rooms. Our approach is based on generating skill-specific interventions and measuring changes in agent predictions. We present a detailed case study analyzing the behavior of a recent agent and then compare multiple agents in terms of skill-specific competency scores. This analysis suggests that biases from training have lasting effects on agent behavior and that existing models are able to ground simple referring expressions. Our comparisons between models show that skill-specific scores correlate with improvements in overall VLN task performance.

%% file: sections/1_introduction.tex
\section{Introduction}
\label{sec:intro}

Following navigation instructions requires coordinating observations and actions in accordance with the natural language. Stopping when told to stop. Turning when told to turn. And appropriately grounding referring expressions when an action is conditioned on some aspect of the environment. All three of these example are required when following the instruction \emph{``Turn left then go down the hallway until you see a desk. Walk towards the desk and then stop.''} In this work, we examine how well current instruction-following agents can execute different types of these sub-behaviors which we will refer to as \emph{skills}.

We situate our study in the popular Vision-and-Language Navigation (VLN) paradigm \cite{anderson2018vision}. In a VLN episode, an agent is spawned in a never-before-seen environment and must navigate to a goal location specified by a natural language navigation instruction. An agent's instruction-following capabilities are typically measured at the episode level -- examining whether an agent reaches near the goal (success rate \cite{anderson2018vision}), how efficiently it does so (SPL \cite{anderson2018evaluation}), or how well its trajectory matches the ground truth path which the human-generated instruction was based on (nDTW \cite{ndtw}). These metrics are useful for comparing agents in aggregate, but take a perspective that has little to say about an agent's fine grained competencies or what sub-instructions it is able to ground appropriately. %

In this work, we design an experimental paradigm based on controlled interventions to analyze fine-grained agent behaviors. We focus our study on an agent's ability to execute unconditional instructions like stopping or turning, as well as, conditional instructions that require more visual grounding like moving towards specified objects and rooms. Our approach leverages annotations from RxR \cite{ku2020room} to produce truncated trajectory-instruction pairs that can then be augmented with an additional skill-specific sub-instruction. We carefully filter these trajectories and generate template-based sub-instructions to build non-trivial intervention episodes that evaluate an agent's ability to ground skill-specific language to the appropriate actions.

To demonstrate the value of this approach, we present a case study analyzing the behavior of a contemporary VLN model \cite{chen2021history}. While we find evidence that the model can reliably ground some skill-specific language, our analysis also reveals that its errors are not random. But rather, they reflect a systematic bias towards forward actions learned during training. For object- or room-seeking skills, we find only modest relationships between instructions and agent actions. Finally, we derive aggregate skill-specific scores and compare across VLN models with different overall task performance. We find that higher skill-specific scores correlate with higher task performance; however, not all skills share the same scale of improvement between weaker and stronger VLN models -- suggesting that improvements in VLN may be fueled by some skills more than others.

\xhdr{Contributions.} To summarize this work, we:
\begin{compactitem}[\hspace{3pt}-]
\item Develop an intervention-based behavioral analysis paradigm for evaluating the behavior of VLN agents,\footnote{\url{https://github.com/Yoark/vln-behave}}\\[-8pt]
\item Provide a case study on a contemporary VLN agent \cite{chen2021history}, evaluating fine-grained competencies and biases, and\\[-8pt]
\item Examine the relationships between  skill-specific metrics and overall VLN task performance.   %
\end{compactitem}

%% file: sections/2_related.tex
\section{Related Work}
\label{sec:related}

\xhdrflat{Vision-and-Language Navigation (VLN).} 
Since its introduction in \cite{anderson2018vision}, many variants of the Vision-and-Language Navigation (VLN) task have been proposed including those in continuous simulators \cite{krantz2020beyond}. We refer the reader to \cite{gu-etal-2022-vision} for a comprehensive survey. In this work, we examine agents in the Room-Across-Room (RxR) dataset \cite{ku2020room} which extends the original VLN task to a multilingual setting with longer, more complex trajectories and pose traces which provide temporal alignment between instruction words and visual observations. There has also been significant modelling work to develop instruction-following agents \cite{Li2022EnveditEE, Guhur2021AirbertIP, Kamath2022ANP, Moudgil2021SOATAS, Hong2021VLNBERTAR,tan2019learning,shen2021much,chen2021history} and we examine three recent models in our analysis \cite{tan2019learning,shen2021much,chen2021history}. 

The RxR task is situated in the \textsc{Matterport3D}\cite{Matterport3D} environments which provide an interface for agents to move through the environment along a graph of panoramic viewpoints taken in real environments. Matterport3D also provides region annotations for room type which we utilize in our experiments. We also leverage annotations from the REVERIE~\cite{qi2020reverie} dataset which extends VLN settings with an additional goal of identifying an object described by a referring expression. Specifically, using the annotations from REVERIEv1 to identify visible objects at each viewpoint.

\xhdr{Evaluating VLN Agents.} 
In standard settings, VLN agents are evaluated by metrics that focus on either the agent reaching the goal efficiently (Success weighted by inverse Path Length \cite{anderson2018evaluation}) or by their trajectory's alignment with the ground truth path (Normalized Dynamic Time Warping \cite{anderson2018vision}). These metrics focus on the agent's performance in aggregate and do not examine agent performance on the level of sub-instruction or skills.

Some works have examined VLN agent behavior more closely by masking or replacing portions of the instructions \cite{zhu-etal-2022-diagnosing,hahnway} and observing the resulting change to overall task metrics like those described above.  Zhu \etal~\cite{zhu-etal-2022-diagnosing} find that VLN agents still achieve relatively high success rates even when all references to visual objects are masked from instructions. These findings cast doubt on the vision-language alignment ability of these agents. \cite{hahnway} also perform masking experiments but come to different conclusions, with some models relying more heavily on nouns than directional words. In both works, agent performance is measured on an episodic level that relies on a sequence of agent decisions; however, single errors in a trajectory may compound and result in misestimating the impact of masked terms. In contrast to these works, we present a  skill-based analysis of VLN agents by constructing specific intervention episodes wherein the appropriate next action is known.

\xhdr{Behavioral Analysis of AI Models.} Recent work in natural language processes has applied behavioral analysis to examine specific skill capabilities. Like us, 
Riberio \etal \cite{Ribeiro2020BeyondAB} develop an intervention paradigm wherein dataset examples are modified in ways such that the desired change in model behavior is knowable. These examples and their associated skills are collected into a ``checklist'' that can be used to evaluate models. Likewise, our work can be construed as generating a checklist of skills for VLN. Yang \etal \cite{Yang2022TestAugAF} follow a similar paradigm and develop a method to automatically generate test cases using a large language models \cite{brown2020language}.

\begin{figure*}[t]
    \centering
    \includegraphics[width=0.97\textwidth]{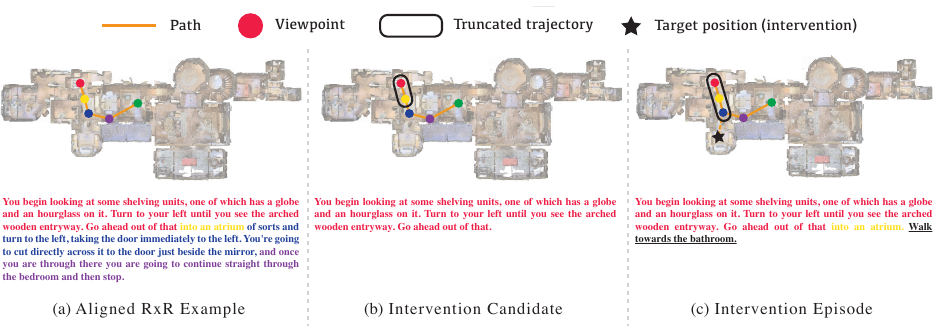}
    \caption{To build skill-specific interventions, we truncate existing RxR episodes based on RxR trajectory-instruction alignments (a$\rightarrow$b). We can then extend the instruction with skill-specific language and identify the next step described by this new instruction (c).}%
    \label{fig:Truncation method}
\end{figure*}

%% file: sections/4_methodology.tex
\section{Analyzing the Behavioral of VLN Agents}
\label{sec:method}

In this work, we examine fine-grained agent competency through the lens of behavioral analysis -- studying how agent decisions change in the presence and absence of skill-specific language instructions. For example, consider object-conditional instructions like ``walk towards the couch''. To demonstrate sensitivity to this instruction, an agent must be more likely to face a couch when presented with this instruction than it would otherwise across a wide range of settings. Note that this analysis examines what agents \emph{do} rather than how they arrive at those decisions and is thus applicable to any VLN agent. To enable this analysis, we develop an intervention strategy that produces trajectories-instruction pairs with and without skill-specific language included. We deploy existing VLN agents on these to examine their decisions and measure their sensitivity to the intervention. The remainder of this section describes our overall methodology -- how intervention episodes are generated and how agents are evaluated  -- and additional skill-specific experimental details are provided in the following section alongside example agent results.

\subsection{Building Intervention Episodes}
\label{subsec:building intervention}

We consider an intervention sample to be a tuple consisting of a trajectory $\tau$, an instruction $I$ that guides an agent to the end of that trajectory,  and an intervention instruction $I_{int}$ that describes some desired skill-specific behavior to be taken from that point. For intervened episodes, an agent will be given the augmented instruction $I + I_{int}$, guided through the trajectory $\tau$ and then its decision will be compared to the expected behavior described in $I_{int}$.  We choose this partial-path construction so we can vary pre-intervention path length while keeping trajectory-instruction alignment similar to standard VLN training settings. \figref{fig:Truncation method}(c) shows one such triplet with a 3-step trajectory, an instruction describing it, and an underlined intervention prompting the agent to move ``towards the bathroom''. We design different interventions to study fine-grained skills in \secref{sec:exp}.

\xhdr{Candidate Instruction-Trajectories Pairs.} To create trajectory-instruction pairs for intervention samples, we leverage the detailed annotations in the RxR dataset \cite{ku2020room}. RxR provides trajectory-instruction pairs along with `pose traces' that provide temporal alignment between instruction words and the annotator's pose in the trajectory. This allows us to associate instruction text segments with nodes along the trajectory -- writing the trajectory as a sequence of node visitations $n_1, n_2, ..., n_T$  with associated sub-instructions $i_1, i_2, ..., i_T$ uttered at each. \figref{fig:Truncation method}(a) demonstrates this by color-coding.
For each RxR example, we form candidate $\tau$, $I$ pairs by truncation -- taking the trajectory up to node $n_j$ ($\tau=\{n_{\leq j}\}$) and instruction text prior to arriving at $n$ ($I_p=\{i_{<j}\}$). This provides a trajectory and the instructions delivering an agent up to the final node.
An example truncation is shown in \figref{fig:Truncation method}(b). For a length $T$ trajectory, this generates $T-2$ candidates; we exclude full-length trajectories because instructions given at penultimate nodes often include explicit directions to stop after moving. Keeping these would add confusion about the appropriate behavior to take after  intervention text is appended.  We consider all English trajectory-instruction pairs from the \texttt{val-unseen-guide} split of RxR~\cite{ku2020room} for this process. %

\xhdr{Filtering and Intervention Instructions.} Not all candidate instruction-trajectory pairs are useful for all interventions -- for example, it may be impossible to "turn left and go forward" for a trajectory that ends on a node without a leftward neighbor. Likewise, an example where a leftward turn is the \emph{only} option could not demonstrate differential agent behavior with and without such an intervention. As described in \secref{sec:exp} below, we filter candidate trajectories for each experiment to ensure agents have both intervention relevant and irrelevant action options in every episode.  

To develop intervention instructions, we manually examined the RxR dataset to identify common skills. In this work, we examine four common skills related to stopping, responding to directional language, and moving relative to objects or room references. For each, we develop a set of language templates to build intervention instructions $I_{int}$. These templates are based on common phrases from RxR instructions and may be conditioned on objects or rooms present in the trajectory. For these object or room references, we leverage the  REVERIEv1 \cite{qi2020reverie} and \textsc{Matterport3D}~\cite{chang2017matterport3d} datasets respectively. \supp{A full list of these templates is provided in the supplementary materials.}

\subsection{Evaluating Agent Sensitivity}
\label{subsec:evaluating}

To evaluate agents on a given intervention episode, we consider both the truncated $(\tau,I)$ and intervened $(\tau,I{+}I_{int})$ trajectory-instruction pairs for that episode. For each, we provide the agent with the instruction and then force it to follow the trajectory $\tau$ until reaching its final node. Matching common teacher-forcing training paradigms, the agent is provided with the observations along the trajectory but its action predictions at each node are ignored in favor of the true next step. At the final node, we then record the agent's predicted action distribution $P(a | \tau, I)$ for each setting where $I$ is either the intervened or truncated instruction. Done over all intervention episodes, this yields a set of distribution pairs $P(a|\tau_j, I_j)$ and $P(a|\tau_j, I_j+I_{int_j})$ which we can use to examine how an agent's beliefs shift under the intervention while everything else about the agent's experience is held constant. We examine the distributions rather than argmax actions to improve sensitivity to shifts in agent belief.

For modern VLN agents, these predicted action distributions correspond to distributions over neighboring viewpoints in the navigation graph as well as a stop action. As such, we can map the desired behavior of most of our interventions to a single or set of neighboring nodes which should have increased or decreased probability under the intervention. For example, telling a skillful agent to ``turn left'' rather than ``go forward'' should shift the predicted probabilities towards neighboring nodes on its left. For some experiments, we introduce additional instruction modification setting to examine the effects of dataset biases in addition to the truncation and interventions constructions. Details are provided in each experimental section.

\xhdr{RxR \vs Intervention Instructions.}  By construction, our generated intervention instructions will tend to contain more short trajectories and instructions than RxR; however, RxR does exhibit a significant variance in path length (see \figref{fig:vln_stats}). On the language side, we use templates that match commonly used phrases in RxR to minimize differences. %

\xhdr{Sample Correlation.} {We note that multiple intervention episodes may be drawn from a single trajectory and many will come from each environment. As such, our measurements may exhibit correlations due to this sampling strategy. To account for these effects when discussing the significance of our results, we apply hierarchical bootstrapping \cite{hboot} when providing confidence intervals and linear mixed effect models when estimating intervention effects. Additional details are provided in the supplementary.}

%% file: sections/5_experiments.tex
\begin{figure}[t]
    \centering
    \includegraphics[width=0.99\columnwidth]{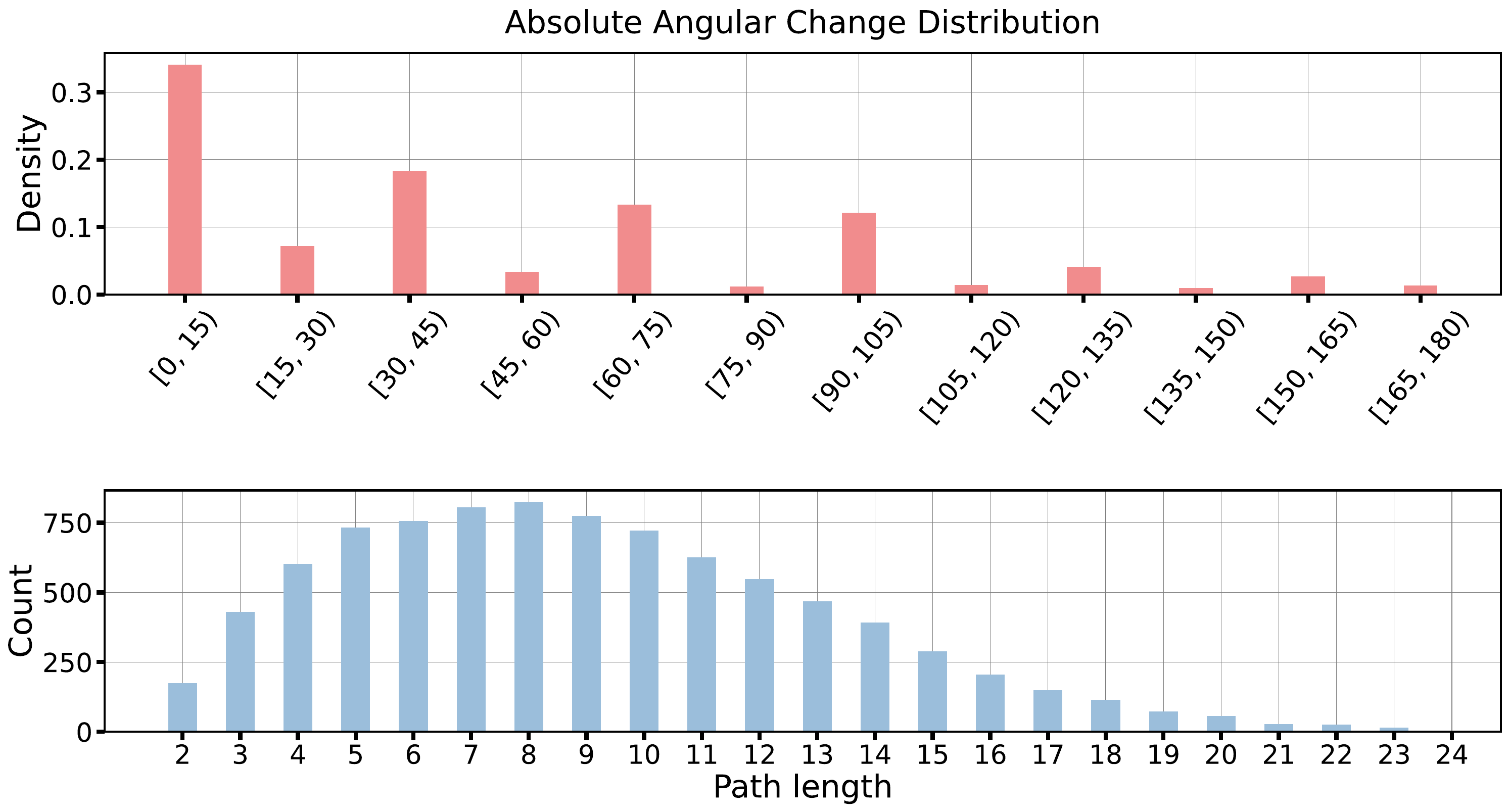}
    \caption{RxR Trajectory Statistics. (Top) Absolute heading change between sequential nodes in training trajectories. There is a strong ``forward'' biases with 34\% of the mass in the 0 to 15$^{\circ}$ bin. (Bottom) The path length distribution of training dataset, showing agents have been exposed to varied length trajectories.}
\label{fig:vln_stats}
\end{figure}

\section{Case Studies on a Recent VLN Agent}
\label{sec:exp}

We instantiate our intervention paradigm for four common skills related to stopping and turning and show the resulting analysis for the recent HAMT \cite{chen2021history} model. HAMT is near state-of-the-art agent for VLN that is based on a multimodal transformer model that jointly encodes trajectory history and instruction text in a hierarchical fashion. HAMT is trained in multiple stages including auxiliary losses and joint imitation and reinforcement learning finetuning. For the experiments below, we use pretrained models and inference code provided by the authors. Like other VLN models, HAMT predicts a distribution over neighboring nodes in the navigation graph and the stop action.

\begin{figure}[t]
    \centering
    \includegraphics[width=0.99\linewidth]{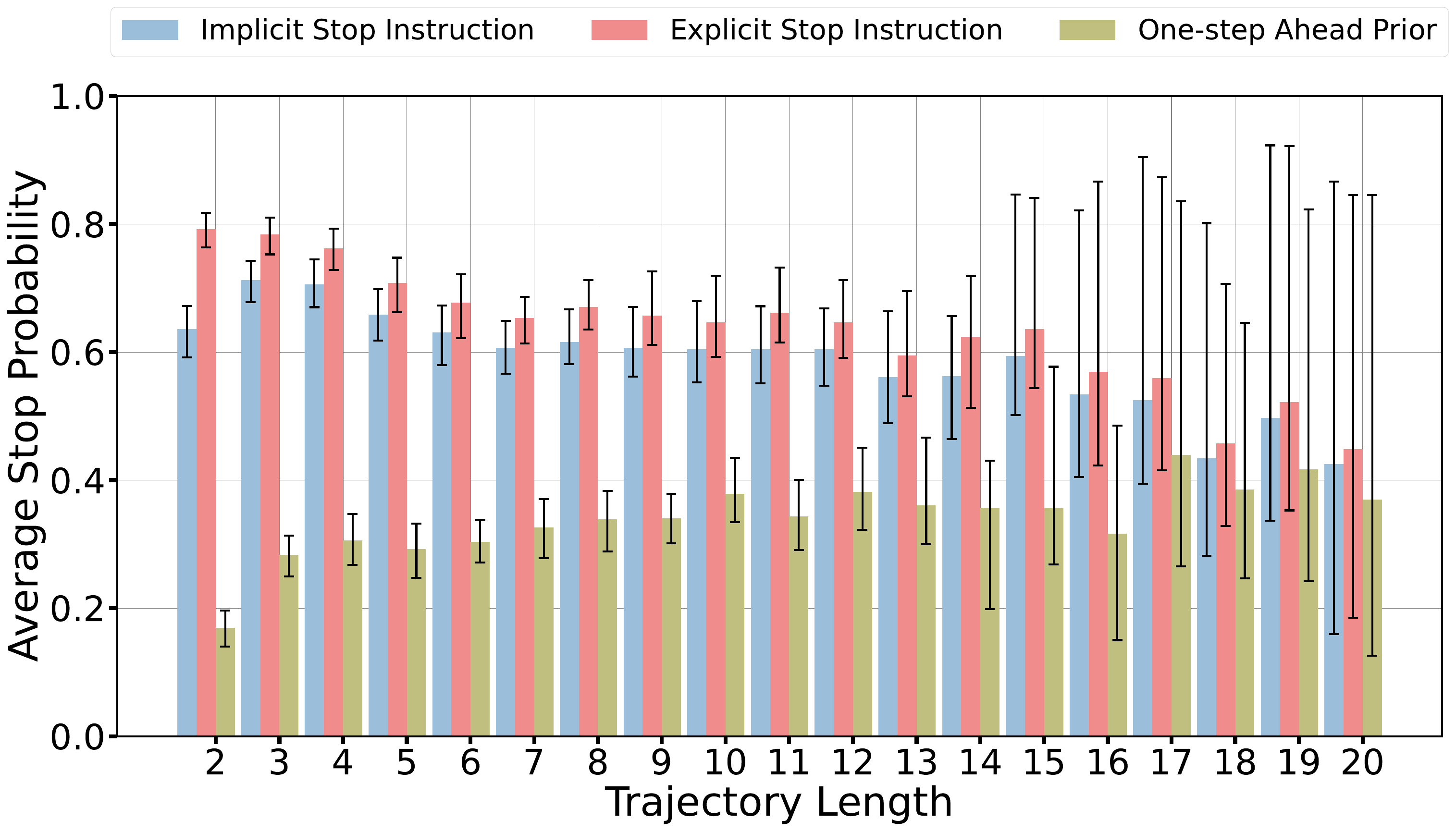}
    \caption{Average Stop Probability vs Trajectory Length for ``implicit stop instruction'', ``explicit stop instruction'' and ``one-step ahead prior''. Agents respond strongly to both stop interventions -- stopping with high probability across all trajectory lengths. Explicit stop instruction produce a stronger effect than implicit.}
    \label{fig:stop}
\end{figure}

\subsection{Stop Instructions}\label{subsec:stop}

To be successful at the VLN task, an agent must declare the \texttt{stop} action within a fixed radius of the goal location described by an instruction. As such, grounding explicit (\myquote{Go to the bedroom and stop.}) and implicit (\myquote{... then go into the bedroom. \texttt{EOS}}) stopping instructions to the \texttt{stop} action is an important skill. In this experiment, we analyze stop behavior for explicit or implicit stop instructions. To assess the effect of path length distributions in RxR, we examine stop behavior across a range of path lengths.

\xhdr{Intervention Details.} All intervention candidates are viable for this experiment as the stop action is always an option and alternative actions (neighboring nodes) always exist. For intervention instructions, we append a short stop instruction such as \myquote{This is your destination.} We note however that the stop experiment offers a complication -- both the truncated  and intervened instructions imply stop actions. The difference being whether this instruction is implicit (truncation) or explicit (intervention). To provide additional comparison with non-stop instructions, we also consider a \emph{one-step ahead} instruction that includes the instruction segment from the terminal node as well (\ie the agent is instructed to make the next step in the trajectory). In total, we produce 8221 intervention episodes. For each episode, we measure the probability of the \texttt{stop} action from agent at the final node of the trajectory.

\xhdr{Results.} \figref{fig:stop} shows average stop probabilities across different trajectory lengths for the truncated implicit stop, intervened explicit stop, and one-step ahead instruction settings. Error bars are 95\% 
hierarchical bootstrap confidence intervals. We find average stop probability to remain fairly constant for implicit and explicit stop instructions.\footnote{Note that longer trajectories have fewer episodes and larger variation.} This suggests agents consistently ground the stop instruction regardless of trajectory length despite biased RxR training trajectory length (see \figref{fig:vln_stats}). 
The plot also suggests stop probability is higher for explicit than implicit stop which are both naturally higher than the one-step ahead setting.

To evaluate statistical significance of the effect, we consider a linear mixed effect model (\texttt{lmer}) where the observed stop probability is assumed to be an effect of the intervention plus random effects from the environment and source trajectory. We find agents have a higher probability of stopping when given explicit rather than implicit stop instructions (0.72 \vs 0.65, effect: 0.07 \texttt{anova}: $p{\approx}0$) and that agents respond to both implicit and explicit stop instructions by increasing stop probability compared to the one-step ahead baseline (effect: 0.36, $p{\approx}0$).

\begin{figure}[t]
    \centering
    \includegraphics[width=1\columnwidth]{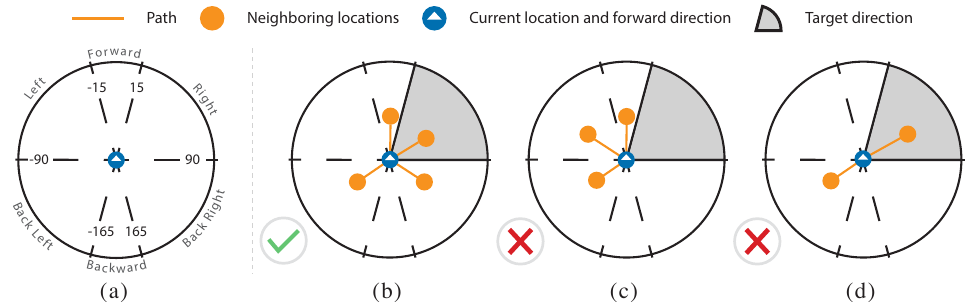}
    \caption{For our directional intervention, we define six directions on the polar axis (a) and establish filters to avoid ungroundable (c) or trivial episodes (d) -- requiring that at a neighboring node is in the target direction and at least two other neighbors are not.}%
\label{fig:direction}
\end{figure}

\begin{figure*}[t]
    \centering
    \includegraphics[width=0.98\linewidth]{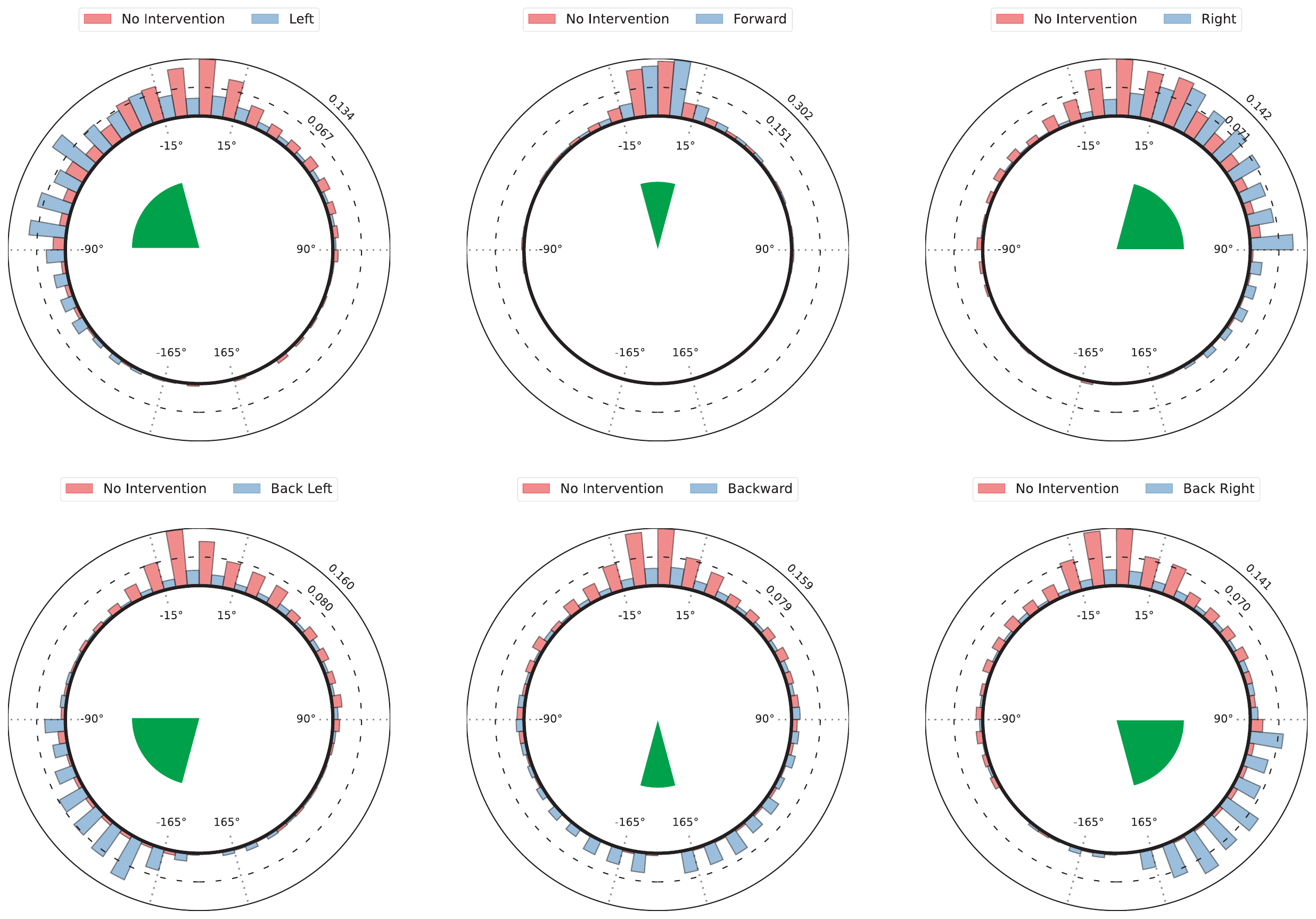}
    \caption{We plot the agent's next step direction probability distribution onto polar axis for easier visualization. We provide results for 6 directions defined in \ref{fig:direction} and contrast between ``No Intervention'' (red) with ``Direction'' (blue). The number on outer circle and middle dotted circle are max and $\frac{\textit{max}}{2}$ respectively. We found the HAMT agent is responsive to all six directional instructions: the probability mass of directional interventions shifts toward the area indicated by directional instructions across all directions comparing to ``No Intervention''}
\label{fig:directionintervention}
\end{figure*}

\xhdr{Summary.} We find the agent responds strongly to implicit and explicit stops across all trajectory lengths and that explicit stop instructions have a stronger effect.

\subsection{Unconditional Directional Instructions}
\label{subsec:direction}

Another foundational skill for following navigation instructions is responding appropriately to directional language. In this experiment, we examine \emph{unconditional directional instructions} which specify directions like \myquote{turn left and go forward} without referencing entities in the environment. This language is frequently used to orient agents in the absence of clear landmarks.
We consider language describing forward/backward motions and turns. Specifically, we explore six direction categories -- forward, backward, left, right, back left, and back right. For each, we define an angular region relative to the agent's heading (canonically 0 degrees) as shown in the top-left of \figref{fig:direction}. During our experiment, we can examine the amount of probability placed on neighboring nodes within these regions.

\xhdr{Intervention Details.} For each direction, we filter intervention candidates to ensure that a) the final node has at least one neighbor in the corresponding direction region and b) that there exist at least two other neighbors outside the direction region. These criteria are demonstrated in \figref{fig:direction}. This ensures that there exists a next step that matches the intervention instruction and that there are multiple alternative actions besides \texttt{stop}. Recall that the agent's action space is to move to a neighbor or to \texttt{stop}, such that turning in place is not possible. So for intervention instructions, we build templates that instruct the agent to face a direction and then go forward (\eg \myquote{Turn right and walk forward}). Early experiments with only direction commands resulted in weaker directional effects. We generate between 3091 and 6745 intervention episodes depending on the direction.

\xhdr{Results.} For each episode, we record the agent's predicted distribution over neighboring nodes. These can be mapped to beliefs over relative angles by associating the probability of visiting neighbor $k$ with the relative angle $\theta_k$ between the agent's heading and neighbor $k$. \figref{fig:directionintervention} shows the distribution of these probabilities over all episodes for each intervention as histograms on polar axes. For convenience, we denote the target direction region with a green arc.

Across all directions, we find the agent either stops (roughly 65\% of the time) or moves  roughly forward in the no intervention setting. For left and right, we see a minor bias towards the corresponding direction despite the agent not receiving any left/right instruction. This reflects a minor structural bias caused by the filtering process. All left (right) episodes include a neighbor to the left (right) and an agent with a bias towards moving roughly forward may select them at a higher rate than other nodes.

For the intervention setting, we see a strong response to directional language. In all cases, the agent stops significantly less (roughly 19\% of the time on average) and we observe a shift in distribution towards the corresponding direction. One outlier is the backwards setting which smears the probability significantly to both the back left and back right. In all settings, some mass still remains in the forward direction -- reflecting the agent's bias towards moving forward learned from the training data as reflected in \figref{fig:vln_stats}.

We accumulate the probability mass into directional bins and evaluate the effect of intervention on the accumulated probability. We again use a linear mixed effect model of the same from as the previous experiment to account for potential correlations in scenes and trajectories. 
 We find the agent exhibits a significantly higher accumulated probability for the corresponding direction with directional instruction than without -- estimating intervention effects as increased accumulated probability for left (0.38, $p\approx0$), right (0.44, $p\approx0$), forward (0.16, $p\approx0$), backward (0.16, $p\approx0$), back left (0.43, $p\approx0$), and back right (0.48, $p\approx0$).

\xhdr{Summary.} We find the HAMT \cite{chen2021history} agent strongly respond to directional language but some dataset biases from training are still evident in a bias towards forward actions.

\subsection{Object-seeking Instructions}
\label{subsec:object}

Beyond directional language, instructions also often use references to nearby objects as convenient landmarks, \eg \myquote{Walk towards the fireplace}. Unlike the language studied in the previous sections, object-seeking instructions require grounding the instruction to the visual scene. We examine simple ``walk towards X'' style object-seeking instructions.

\xhdr{Intervention Details.} We leverage the object annotations from the REVERIE \cite{qi2020reverie} dataset to build intervention episodes. Specifically, we retain an intervention candidate if a REVERIE object is visible from its terminal node, the object is no more than 3m away, there exists a neighboring node with a heading that is within 15 degrees of the object's heading, and there at least two neighbors. That is to say, trajectories that end near a visible object and a non-trivial navigation action can {reasonably} move towards it. We exclude common structural objects like doors, windows, shelving, railings, etc.~as it is often unclear which of multiple occurrences an agent should move towards. For instructions intervention instructions, we append the template \myquote{Walk towards the [object]} where \texttt{object} is the object name from REVERIE.  In total, we generate 839 intervention episodes targeting the following objects in decreasing order of occurrence: chair, table, picture, cushion, curtain, plant, cabinet, gym equipment, stool, chest of drawers, bed, towel, bathtub, tv monitor, and seating.

\begin{figure}[t]
    \centering
    \includegraphics[width=0.99\linewidth]{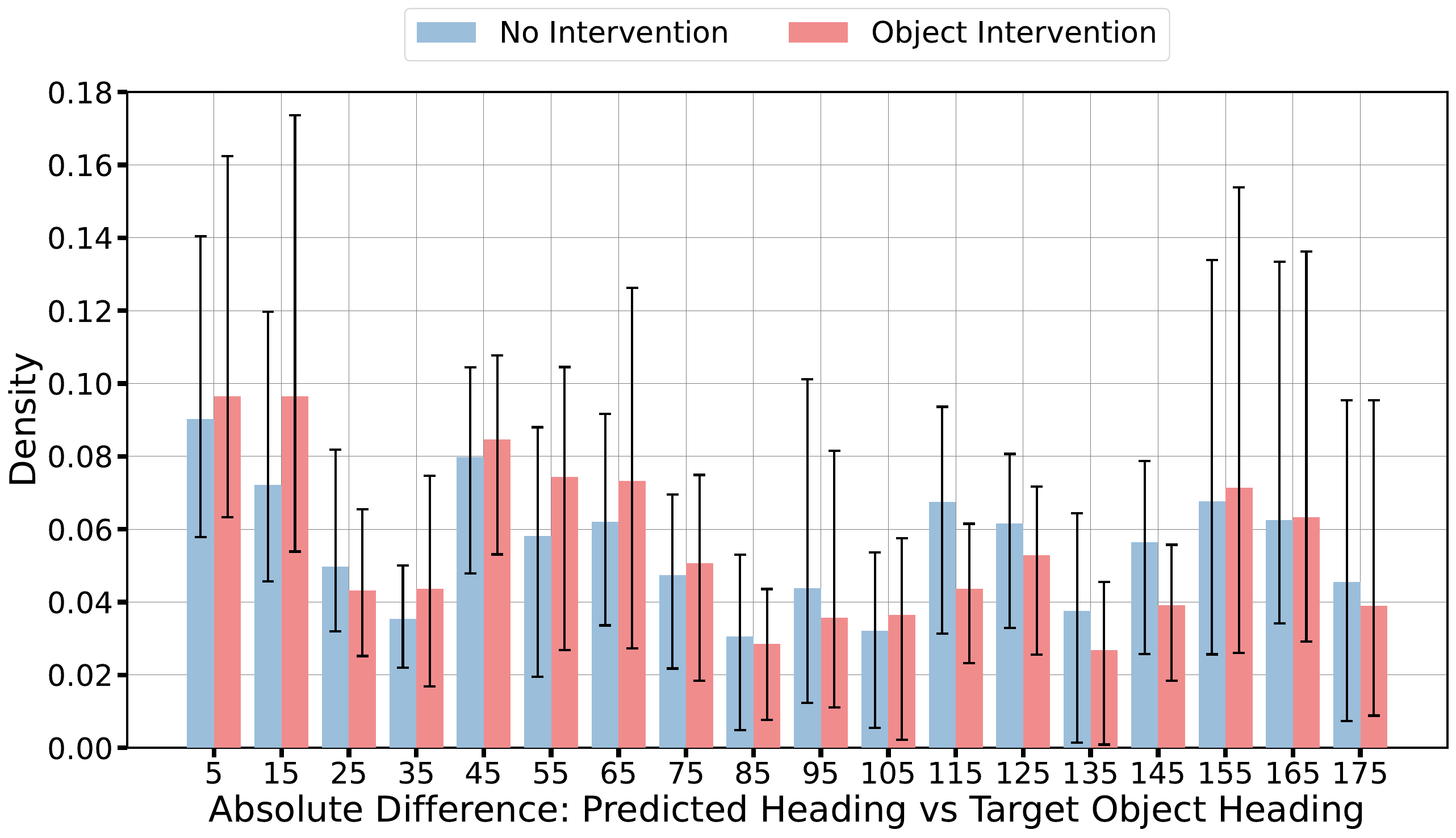}
    \tightcaption
    \caption{The distribution of absolute difference between model prediction and target object direction for intervention and no intervention settings.}
    \label{fig:diff_object_pred}
\end{figure}

\xhdr{Results.} For each episode, we record the agent's predicted distribution over neighboring nodes at the terminal node. We map these to a distribution over absolute angular errors relative to the object. For each neighbor $k$, we compute the difference in heading angle between node $k$ and the object. We can then associate the probability of visiting neighbor $k$ with an angular distance to the target object. These probabilities are accumulated and normalized to produce \figref{fig:diff_object_pred} which presents distributions over angular distance for the intervention and no-intervention settings.  

This intervention shows a weak effect -- with the agent reducing angular error in the presence of the intervention instruction somewhat (blue \vs red bars in \figref{fig:diff_object_pred}). We again leverage a linear mixed effect model to evaluate the effect of intervention on the accumulated probability within 15 degree of absolute angular difference. We find weak fixed effect of 0.069 (\texttt{anova}, $p\approx0$) for intervention \vs non-intervention. However, both settings exhibit a wide range of angular errors suggesting that target objects are not being grounded accurately -- recall that all trajectories have neighboring nodes that would incur no more than 15 degrees of error. To explore this further, we also examine a baseline \texttt{Forward Bias} agent that places probability on neighbors inversely proportional to their relative heading. We find this baseline exhibits a similarly shaped error distribution -- suggesting the agent may be taking forward actions when uncertain about the target object.
As in our other experiments, the no intervention setting is more likely to stop than the intervention (65\% \vs 37\%).

\xhdr{Summary.} We find evidence for only a weak tendency to move towards referenced objects for this agent.

\subsection{Room-seeking Instructions}
\label{subsec:room}

Agents may also be asked to navigate to specific rooms, often without specific directional language describing how to access them -- \eg \myquote{go to the kitchen.} In this experiment, we examine these \emph{room-seeking} instructions both in the setting where the room is likely visible and when an agent may need to search for it nearby. We note that this latter task is beyond the scope of standard VLN instructions and examine it as a test of generalization. Below, we denote the case where a room is likely visible as a 1-hop setting and extensions beyond this as k-hop.

\xhdr{Intervention Details.} Using  \textsc{Matterport3D}~\cite{chang2017matterport3d} room region annotations, we associate each node in all trajectories with a room label. For 1-hop settings, we retain intervention candidates where the terminal node has at least one neighboring node with a different room type. For k-hop, we extend this to consider neighbors that are k steps away from the current node.
For intervention instructions, we append a template \myquote{Walk towards the [room].} where \texttt{room} is replaced with the corresponding room name from \textsc{Matterport3D}. As k-hop episodes involve agents making multiple decisions, we also append \myquote{This is your destination.} afterwards to encourage agents to stop once reaching the room. We generate 8614 intervention episodes for 1-hop setting, and 17204 to 27454 episodes for n-hop settings.

\begin{figure}[t]
    \centering
    \includegraphics[width=0.99\linewidth]{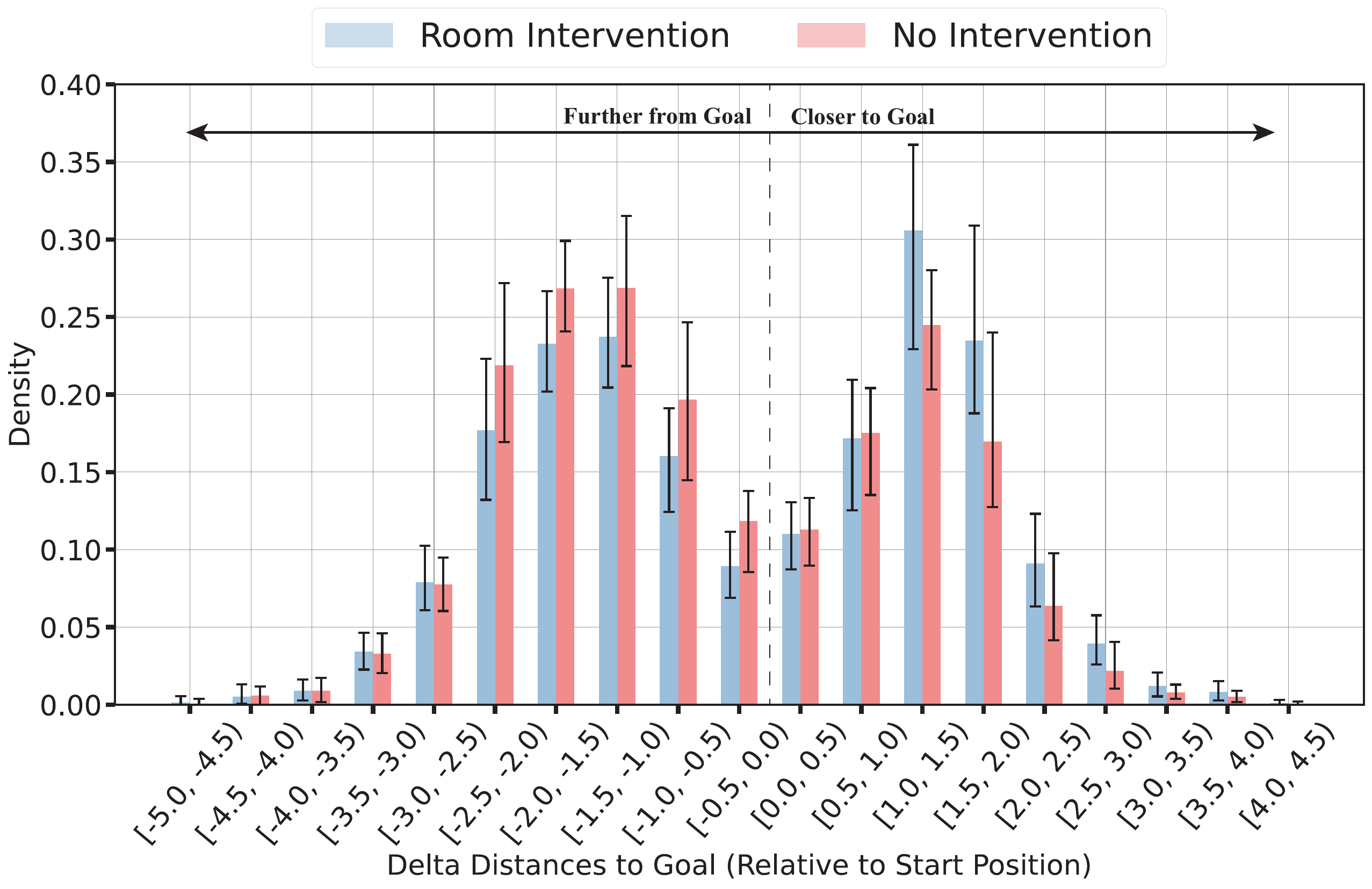}
    \tightcaption
    \caption{Distribution of delta distance to target room type. The delta distance difference of distance to target room (relative to start position) with or without intervention. Positive delta distance means the agent move closer to room of target type with intervention than otherwise. The distribution shift towards right with intervention than otherwise, indicates the agent is responsive to room-seeking instruction. (-0.15 \vs -0.41, $p\approx 0$ )}
    \label{fig:1hop}
\end{figure}

\xhdr{1-Hop Results.} For each episode, we record the agent's predicted distribution over neighboring nodes at end of the trajectory. To measure agent progress towards nodes with target room type, we map these to a distribution over change in geodesic distances to the \emph{nearest} node with the target room type. This is done analogously to previous sections such that the probability of visiting node $j$ from the final trajectory node $T$ is associated with the delta geodesic distance $\Delta d = d_{geo}(n_T, n_{near}) - d_{geo}(n_j, n_{near})$ to the nearest node with the target room type $n_{near}$. The probabilities are accumulated and shown in \figref{fig:1hop} -- values greater than zero represent the agent moving \emph{closer} to nodes with the target room type. We observe a right-shift in the density suggesting the agent responds somewhat to the intervention. Again using a linear mixed effect model, we estimate the effect of intervention on the delta geodesic distance as 0.26 (\texttt{anova}, $p\approx0$) for intervention vs no intervention. However, the agent does not reliably place strong beliefs on neighbors with the target room type -- negative median delta distance and significant mass to the left of zero.

\begin{figure}[t]
    \centering
    \includegraphics[width=0.99\linewidth]{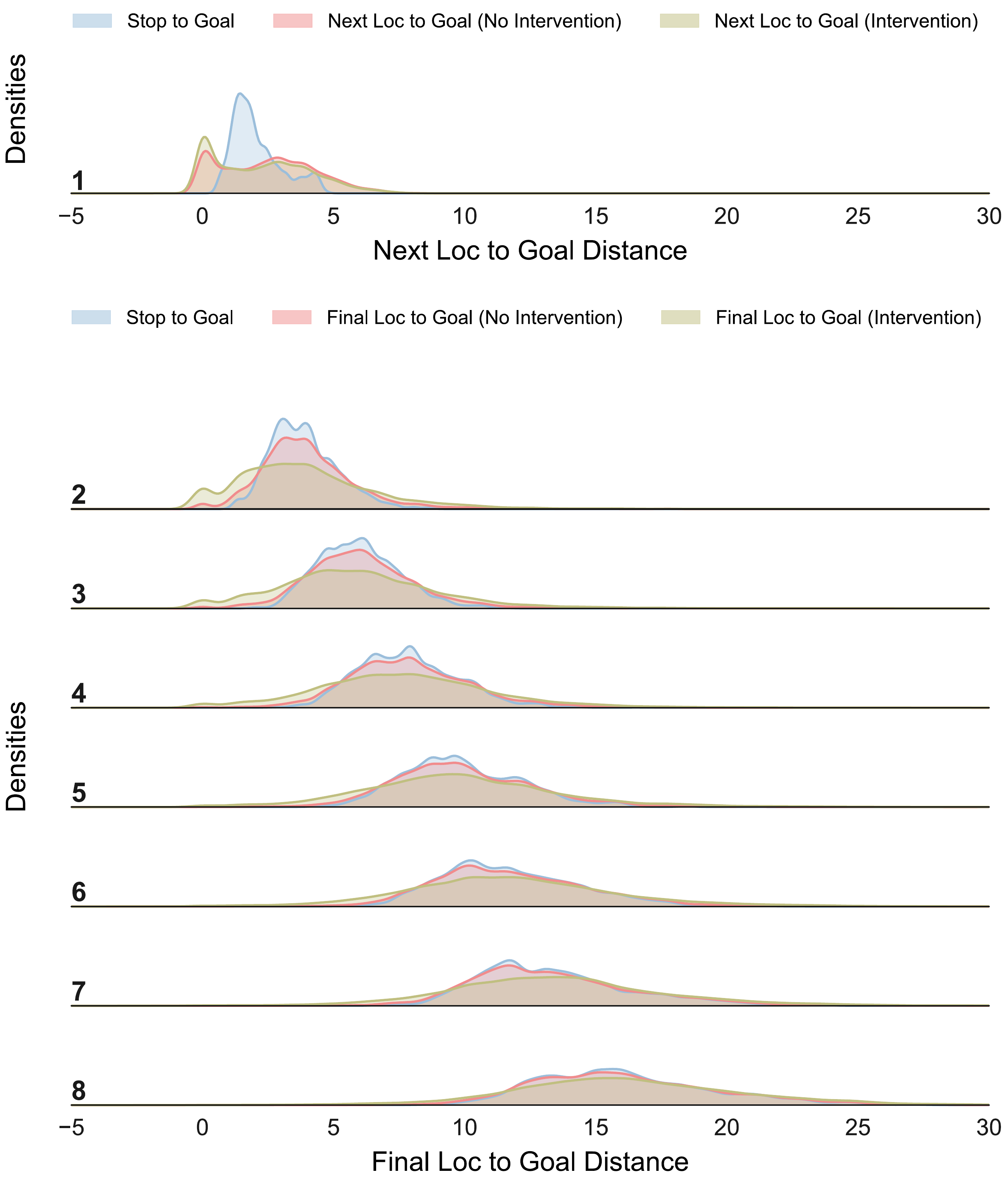}
    \tightcaption
    \caption{Distribution of geodesic distance to nearest target room location for k-hop room-seeking experiments. \texttt{Stop to Goal} is a baseline agent that always takes the stop action.}
    \label{fig:ridgeplot_room}
\end{figure}

\xhdr{k-Hop Results.} For each k-hop episode, we force the agent to follow the trajectory until it ends at node $n_T$ and then execute the agent by taking argmax actions until \texttt{stop} is called and then record the final position $n_{end}$. We report the distance to the nearest node with target room type from here, $d_{geo}(n_{end}, n_{near})$.
We shows a ridgeline plot in \figref{fig:ridgeplot_room} comparing these distributions for 1- to 8-hops. %

 We find error increases with target room distance. We again leverage a linear mixed effect model to evaluate the effect of intervention on $d_{geo}(n_{end}, n_{near})$. We find weak fixed effect of $\leq-0.1$ for intervention \vs non-intervention for 3,4, and 5 hops with 95\% confidence). Note that sample size varies with number of hops. Overall, this suggests agents have limited ability to search for rooms based on common sense exploration -- perhaps unsurprising given that RxR instructions typically provide step-by-step guidance.

\xhdr{Summary.} The HAMT \cite{chen2021history} agent is weakly sensitive to room type reference instructions when the room is visible (within one hop) but lacks the ability (room type instruction have weak or none effects) to perform common sense exploration to find further away rooms (k-hop). Overall sensitivity is low, suggesting the agent may not rely on room-specifying portions of instructions when navigating.

\section{Comparing Sensitivity Across Agents}
\label{sec:scoring}

For each skill-specific intervention, we can identify a set of neighbor actions (neighboring nodes or \texttt{stop}) that correspond to a correct grounding of the intervention instruction -- matching the filtering criteria used in their constructions. For stop instructions, this is the stop action. For turn instructions, neighbors within the corresponding direction angle range are valid. For object-seeking, neighbors within 15$^\circ$ of the object heading. For room-seeking, neighbors with target room type.

To compare across VLN models, we examine the average probability mass they place on these correct actions. Denoting the set of correct actions as $\mathcal{N}_e$ for an intervention episode $e$, we can write a scoring function over a set of intervention episodes $\mathcal{E}_s$ for skill $s$  for an agent $f$ as
\begin{equation} Score(f,s) =\frac{1}{|\mathcal{E}_s|}\sum_{e\in\mathcal{E}_s} \sum_{a_n\in\mathcal{N}_e} P_f(a_n | \tau_e, I_e+I_{int_e})  
\end{equation}
where $P_f(a_n | \cdots)$ is agent $f$'s predicted probability for action $a_n$ at the end of the intervention trajectory. Higher scores reflect greater certainty in selecting a grounded action on average across all episodes.

\tabref{tab:scores} shows these scores for three VLN agents of varying RxR task performance (EnvDrop $<$ EnvDrop (ViL CLIP) $<$ HAMT). We find that improved model performance on the overall RxR task tends to also leads to improvements on skill-specific scores. However, improvements are not uniform across the skills and agents are more proficient at stopping and turning instructions than those referencing objects or rooms.

\input{sections/score_table}

%% file: sections/score_table.tex
\begin{table}[t]
\centering
\renewcommand*{\arraystretch}{1.5}
\setlength{\tabcolsep}{4pt}
\resizebox{\columnwidth}{!}{
\begin{tabular}{l p{0.3em} c c c c p{0.3em} s}
    \toprule
    Method && Stop & Turn & Object & Room && Avg.\\
    \midrule  
    EnvDrop \cite{tan2019learning} && 62.65 & 27.14 & 11.06 & 23.64 && 31.12 \\
    EnvDrop (ViL CLIP) \cite{shen2021much} && 66.76 & 27.45 & 12.83 & 26.82 && 33.47 \\
    HAMT \cite{chen2021history} && 71.65 & 43.74 & 12.00 & 26.63 && 38.50 \\
     \bottomrule
\end{tabular}}
\caption{We report scores for each skill type for VLN models with varying RxR task performance (EnvDrop ${<}$ EnvDrop (ViL CLIP) ${<}$ HAMT). We find individual skill performance tends to improve with overall task performance, but not equally over all skills. Object- and room-seeking instructions require further study.}
\label{tab:scores}
\end{table}

%% file: sections/6_discussion.tex
\section{Discussion}
\label{sec:disc}

In this work, we introduced an analysis paradigm for studying fine-grained skill competency in VLN agents. To show its value, we presented a case study on a recent VLN model. This provided insights into agent behavior including significant differences in performance on unconditional (stop and turn) and conditional (object- and room-seeking) skills. Finally, we presented a comparison between models in terms of skill-specific scores.

Skill-specific analysis like we present here can provide insight into the types of things we can reasonable expect from our models and those which will require further study. Building explicit test cases for desired skills can serve as ``unit tests'' on our path to more complex instruction following systems. This work is a step in that direction.

%% file: supplementary/discussion.tex
\section{Discussion}
We discuss the insights regarding model behaviour, as well as some future directions. Main paper's goal is to develop a framework for skill-based behavioral analysis, despite that, we could provide some speculation on the underlying reason of model behaviour in the hope of benefiting the future development of embodied agents.

\vspace{2pt}\noindent--\underline{Low room/object sensitivity of HAMT.} 
Low sensitivity of object and room seeking implies a weakness of the agent in spatial relation reasoning and vision-language alignment. We suspect this resulted from lack of specific proxy tasks, and visual features only capturing limited information (as also stated in [22]). We encourage people to design specific architectures, build proxy tasks addressing spatial relation reasoning, and incorporate richer object information or object representation learning modules.

\vspace{2pt}\noindent-- \underline{HAMT vs.~EnvDrop (Stop \& Turn).}
Architecture difference (HAMT vs EnvDrop: Transformer vs Recurrent Neural Network) might give HAMT an advantage in both Stop and Turn. Further for Turn, we believe some proxy training tasks unique to HAMT brought the advantage. We suspect Single-Step Action Regression, and Spatial Relationship Prediction are helpful. Former predicts action heading and elevation directly from given instruction, history, and current observation; latter predicts relative spatial position of two views given visual feature angle feature or both. Further analysis could be interesting future work.
 
\vspace{2pt}\noindent-- \underline{EnvDrop vs.~ EnvDrop (CLIP).}
CLIP may provide improved semantics, but not action-grounding benefits. A full-scale  component-wise analysis is out-of-scope for this paper, but would be an interesting application of our behavioral analysis framework that our code release could support in the future.

%% file: supplementary/stats.tex
\section{Data Correlation Analysis}
Our dataset represents a finite, correlated sample from the space of all instruction-trajectories pairs in indoor scenes. There may be correlation within trajectories from the same scan or from interventions drawn from the same trajectory. We conducted Hierarchical bootstrapping and linear mixed effect modeling to account for the correlation in data.

\vspace{2pt}\noindent -- \underline{Hierarchical bootstrapping for CIs.} We use hierarchical bootstrap resampling~\cite{hboot}
(scenes$\rightarrow$trajectories) to correctly simulate a new draw from the underlying population we are studying. Then we obtain confidence intervals from the new draw.

\vspace{2pt}\noindent -- \underline{Linear mixed effect modeling.} We model each of our interventions with a linear mixed effect model where each scan and trajectory are modeled as imparting a random slope and intercept along with a overall fixed intervention effect -- i.e.~modeling the effect for an episode $i$ taken from scan $j$ and trajectory $k$ as 
\begin{equation*}
    \mbox{effect}_i {=} \left(w_{\mbox{fix}}+w_{\mbox{scan}_j}+w_{\mbox{traj}_k}\right)*I_i + b_{\mbox{fix}} + b_{\mbox{scan}_j} + b_{\mbox{traj}_k} 
\end{equation*}
where $w_{\mbox{scan}_j}, w_{\mbox{traj}_k}, b_{\mbox{scan}_j},$ and $b_{\mbox{traj}_k}$ are modeled as random effects and $I_i$ is a binary variable indicating whether this episode contains an intervention. Models were fit using \texttt{lmer} in \texttt{R} and significance of fixed effects were evaluated through the \texttt{anova} command. We provided analysis for HAMT, ENVDROP-IMAGENET, ENVDROP-CLIP in main paper and appendix.

%% file: supplementary/results_extended.tex
\begin{figure}[t]
    \centering
    \includegraphics[width=1\columnwidth]{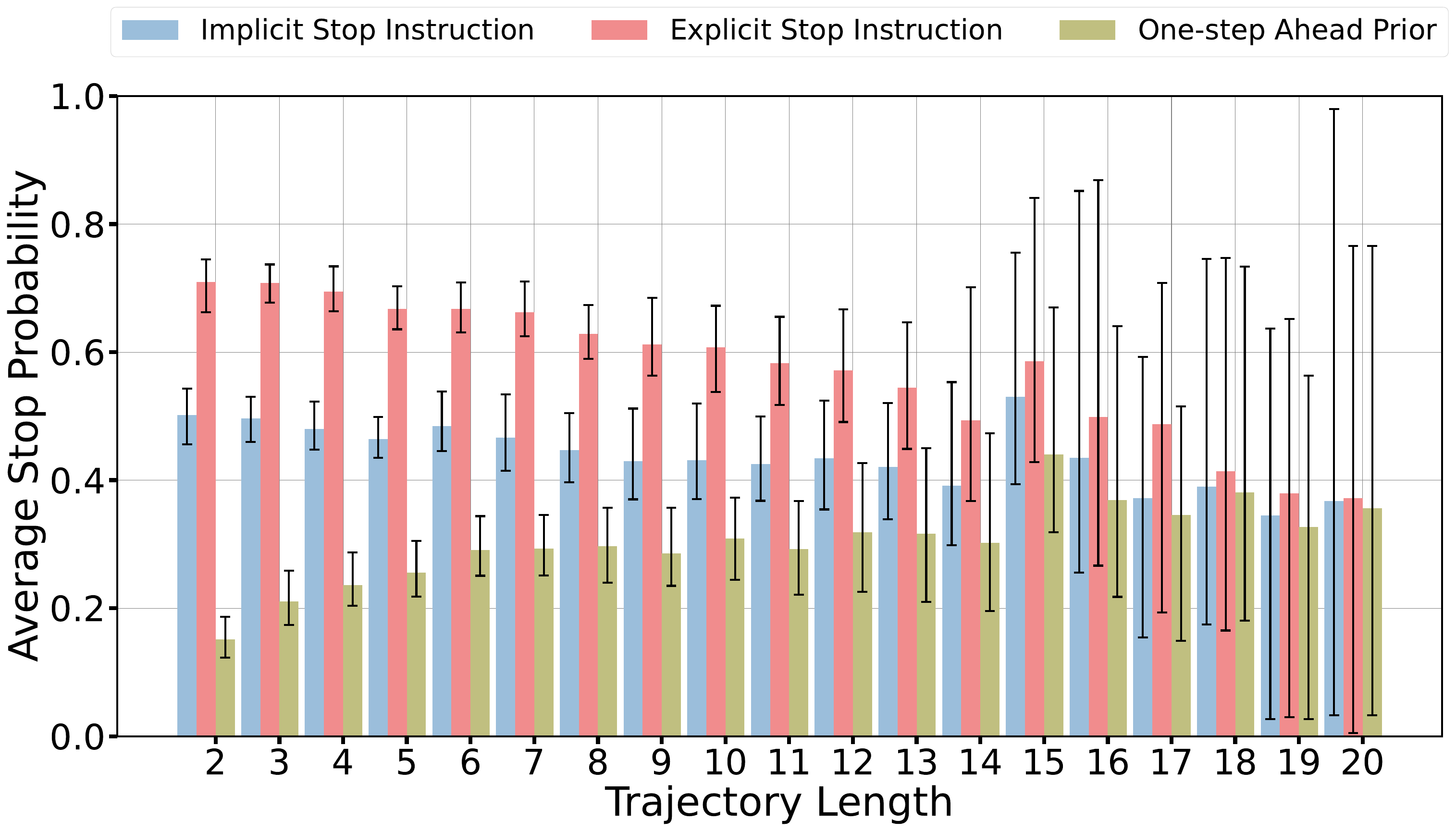}
    \caption{(Envdrop--clip) Average Stop Probability vs Trajectory length instruction for ``implicit stop instruction'', ``explicit stop instruction'' and ``one-step ahead prior''. We find agents respond strongly to both implicit and explicit stop interventions at earlier steps -- stopping with high probability across shorter trajectory lengths. (Until around sixteen) Explicit stop instructions produce a stronger effect than implicit.}
\label{fig:stop_clip}
\end{figure}

\begin{figure}[t]
    \centering
    \includegraphics[width=1\columnwidth]{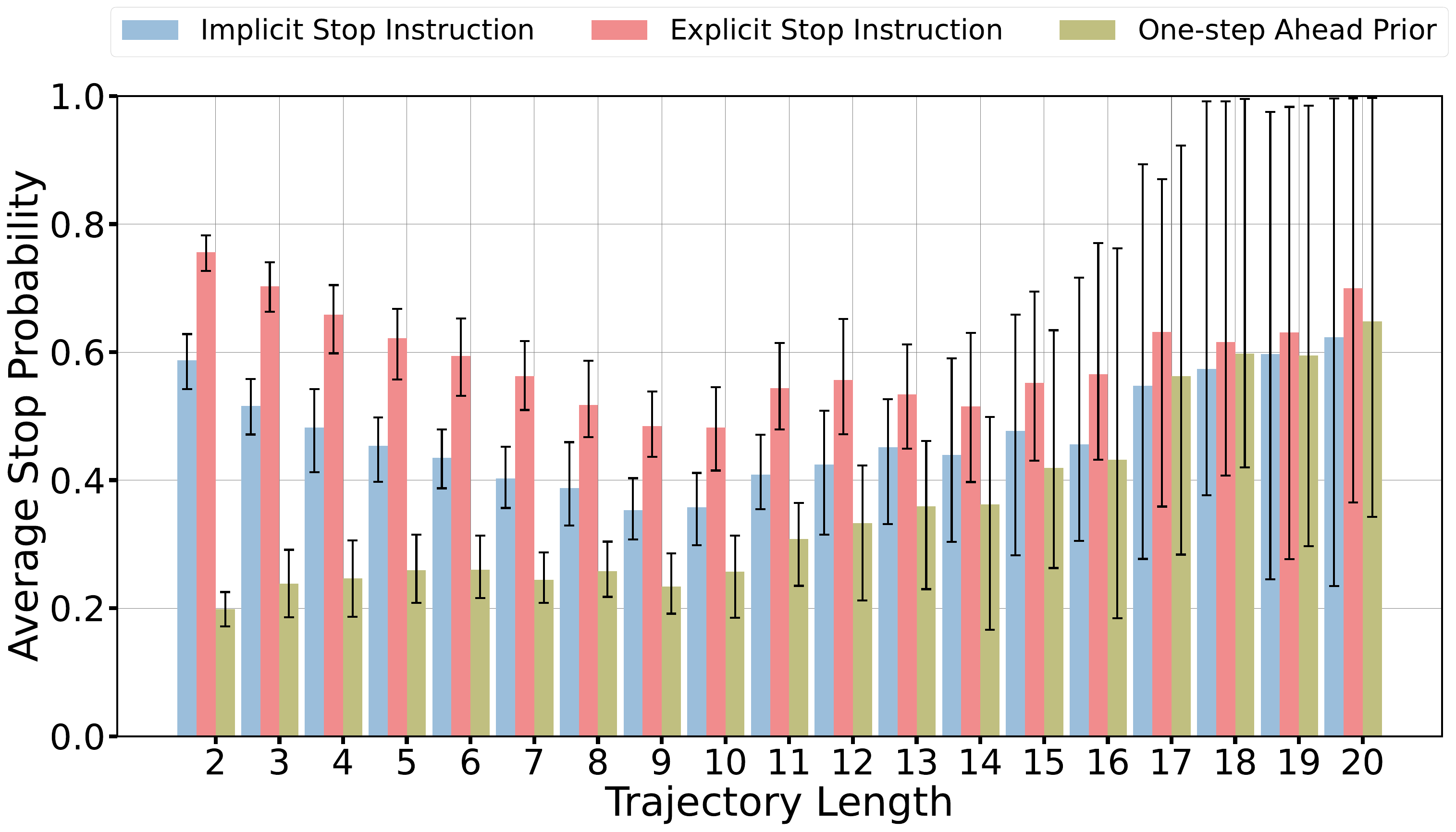}
    \caption{(Envdrop--imagenet)Average Stop Probability vs Trajectory length instruction for ``implicit stop instruction'', ``explicit stop instruction'' and ``one-step ahead prior''. We find agents respond strongly to both implicit and explicit stop interventions -- stopping with high probability across all trajectory lengths. Explicit stop instructions produce a stronger effect than implicit.}
\label{fig:stop_imagenet}
\end{figure}

\section{Additional Case Studies for Envdrop-clip and Envdrop-imagenet}
We provide complete analysis for the two additional VLN agents we tested: \textsc{Envdrop--clip} and \textsc{Envdrop--imagenet}. These were not included as case studies in the main paper due to space.

\subsection{Stop}

\figref{fig:stop_clip} and \figref{fig:stop_imagenet} show average stop probabilities across different trajectory lengths for the truncated implicit stop, intervened explicit stop, and one-step ahead instruction settings for \textsc{Envdrop--imagenet} and \textsc{Envdrop--clip}. Error bars are 95\% hierarchical bootstrap confidence intervals. For \textsc{Envdrop--clip}, we find the average stop probability to remain fairly constant for shorter trajectory lengths (until around sixteen) under both implicit and explicit stop instructions, but dropped at longer trajectory lengths (from sixteen to twenty). This suggests agents ground the stop instruction better for shorter trajectories. And the plot also suggests stop probability is higher for explicit than implicit stop and both are higher than one-step ahead setting. 

To evaluate statistical significance of the effect, we again use \texttt{lmer} where the observed stop probability is assumed to be an effect of the intervention plus random effects from the environment and source trajectory. We find agents have a higher probability of stopping when given explicit rather than implicit stop instructions (0.66 \vs 0.47, effect 0.19 \texttt{anova:} $p\approx0$), and the agent responds to both implicit and explicit stop instructions by increasing stop probability compared to the one-step ahead baseline (effect 0.22, $p\approx0$). For \textsc{Envdrop--imagenet}, we find the average stop probability to remain fairly constant for implicit and explicit stop instructions across all trajectory lengths. This suggests the agent can ground to both implicit and explicit stop instructions regardless of trajectory length. The stop probability for explicit stop instruction is higher than implicit stop instruction, and both are higher than one-step ahead setting. We find agents have a higher probability of stopping with explicit rather than implicit stop instructions (0.63 \vs 0.47, effect: 0.16 $p\approx0$), and the agents respond to both implicit and explicit stop instructions by increasing stop probability compared to the one-step ahead setting (effect 0.22, $p\approx0$)

Note \textsc{Envdrop--imagenet} has a tendency to stop more likely for longer trajectory than \textsc{Envdrop--clip}. This might suggest the correlation between trajectory lengths and stop probability for \textsc{Envdrop--imagenet} is stronger. 

\xhdr{Summary.} We find both \textsc{Envdrop--imagenet} and \textsc{Envdrop--clip} respond strongly to implicit and explicit stops across most trajectory lengths and explicit stop instructions have a stronger effect. In addition, we find \textsc{Envdrop--clip} tends to have a lower probability of stopping at longer trajectories regardless of stop instructions.
\subsection{Unconditional Directional Instructions}

\begin{figure*}[t]
    \centering
    \includegraphics[width=\linewidth]{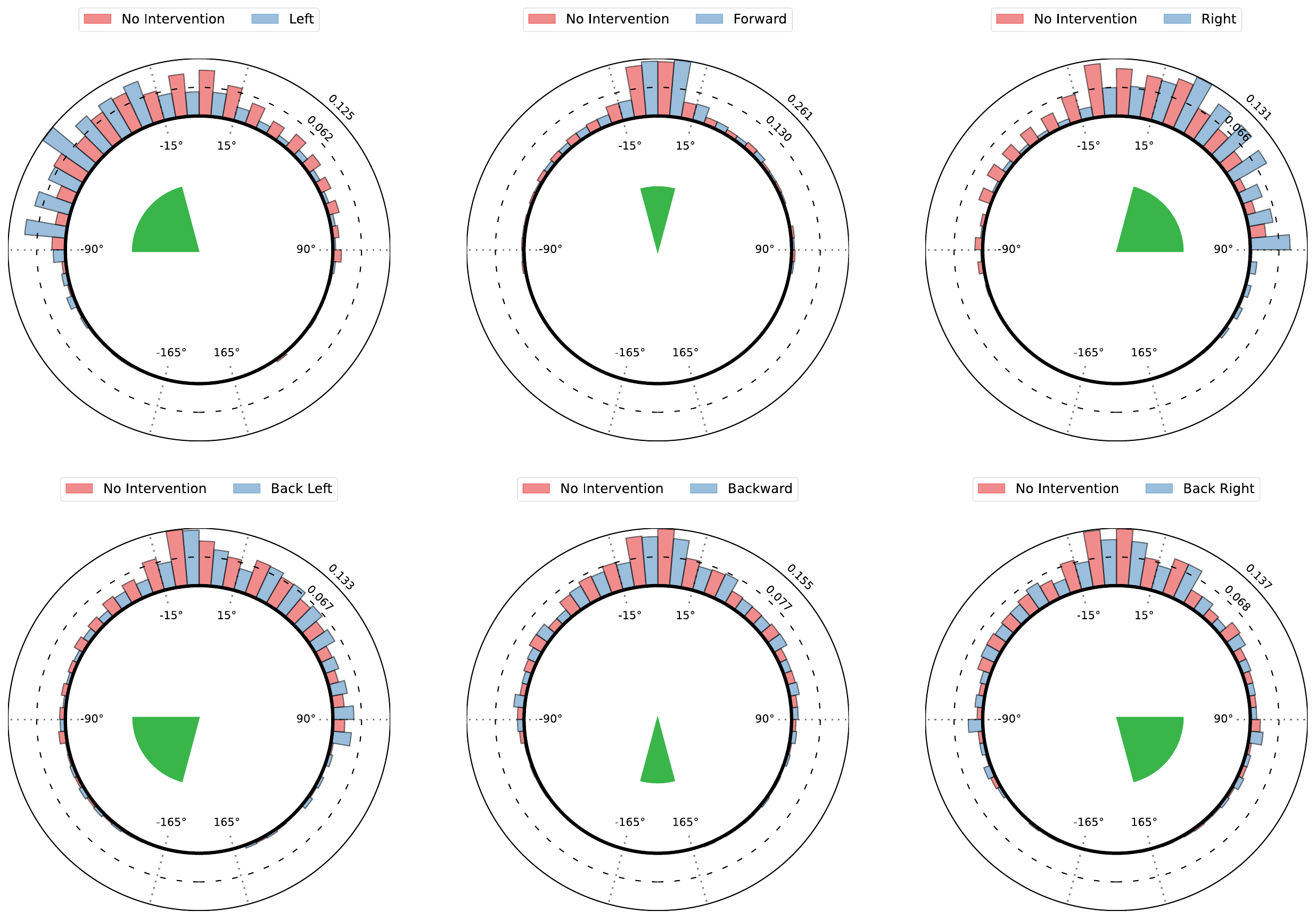}
    \caption{(\textsc{Envdrop--clip}) We plot the next step direction probability distribution of the agent onto polar axis for easier visualization. We provide results for six directions and contrast between ``No Intervention'' (red) with ``Direction'' (blue). The number on the outer circle and middle dotted circle are max and $\frac{\textit{max}}{2}$ respectively. We found the \textsc{Envdrop--clip} agent is responsive to only three directional instructions: forward, left and right. The probability mass of directional interventions shifts toward the area indicated by those three directional instructions compared to ``No Intervention''.}
\label{fig:directionintervention_clip}
\end{figure*}

\begin{figure*}[t]
    \centering
    \includegraphics[width=\linewidth]{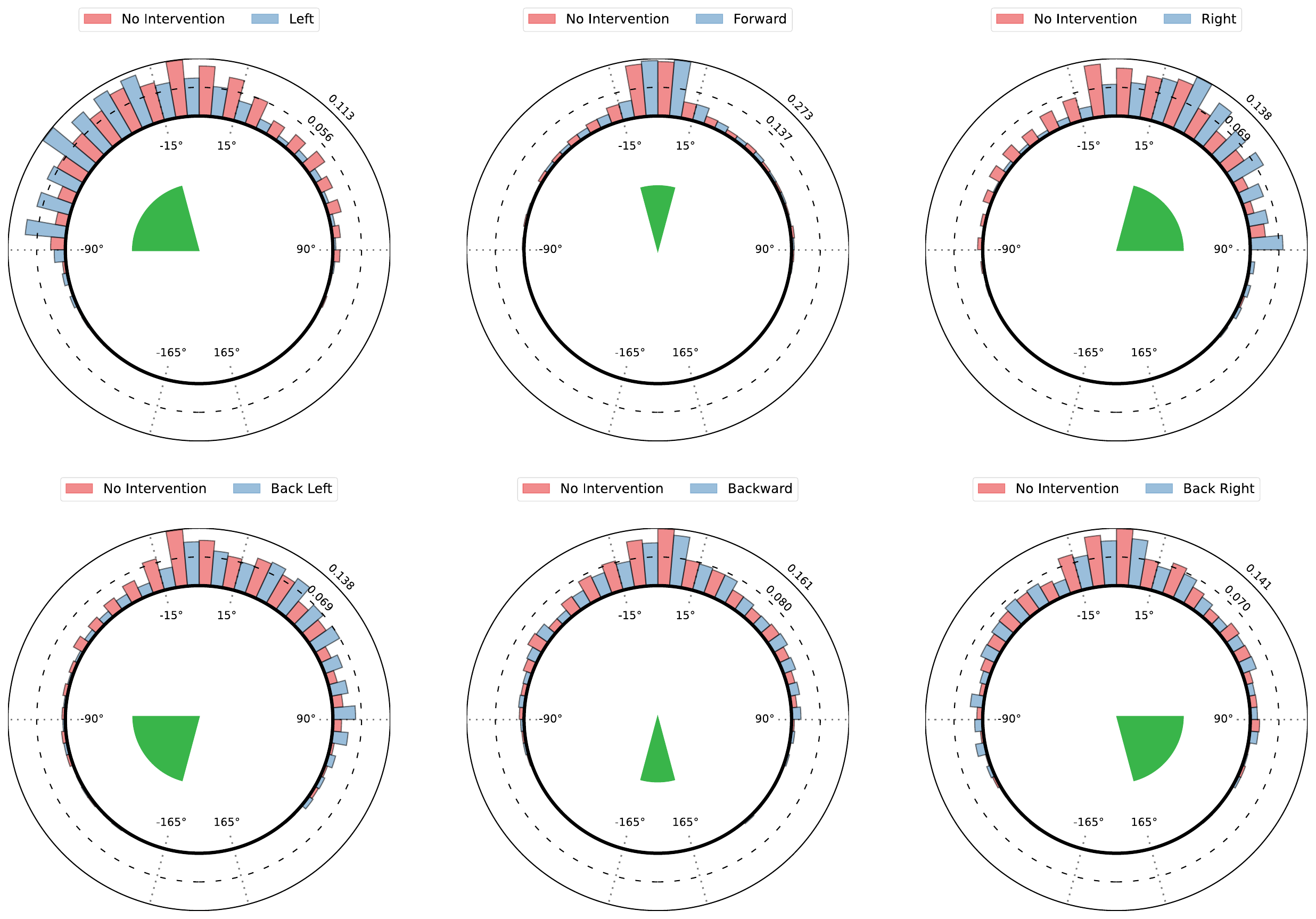}
    \caption{(\textsc{Envdrop--imagenet}) We plot the next step direction probability distribution of the agent onto polar axis for easier visualization. We provide results for six directions and contrast between ``No Intervention'' (red) with ``Direction'' (blue). The number on the outer circle and middle dotted circle are max and $\frac{\textit{max}}{2}$ respectively. We found the \textsc{Envdrop--imagenet} agent is responsive to only three directional instructions: forward, left and right. The probability mass of directional interventions shifts toward the area indicated by those three directional instructions compared to ``No Intervention''.}
\label{fig:directionintervention_imagenet}
\end{figure*}

\figref{fig:directionintervention_clip} and \figref{fig:directionintervention_imagenet} show the distribution of probabilities over all episodes for each directional intervention as histograms on polar axes. For convenience, we denote the target direction region with a green arc at the center of each plot.

For \textsc{Envdrop--clip}, across all directions, we find the agent either stops (roughly 46\% of the time on average) or moves in a roughly forward direction in the no intervention setting. There is a slight bias towards left or right in those settings. However, the agent does not receive any left/right instruction, so this reflects a minor structural bias caused by the filtering process. All left (right) episodes include a neighbor to the left (right) and an agent with a bias towards moving roughly forward may select them at a higher rate than nodes in the backward direction.

For the intervention setting, we see a strong response to directional language for forward, left and right. For these three settings, the agent stops significantly less (roughly 21\% of the time on average) and we observe a shift in distribution towards the corresponding direction. Similarly as before, we accumulate the probability mass into directional bins and evaluate the effect of intervention on the accumulated probability. We again use \texttt{lmer} the same as before to account for potential correlations in scenes and trajectories. We find the agent exhibits a significantly higher accumulated probability for forward, left, and right direction with directional instruction than without -- estimating intervention effects as increased accumulated probability for forward (0.08, $p\approx0$), left (0.36, $p\approx0$), right (0.34, $p\approx0$)

For backward, back left, and back right, the agent does not have a good response to directional language. We find the agent either stops (roughly 58\% of the time on average), moves forward (reflecting forward bias the agent learned during training), or responds to part of the instruction. We created backward, back left, and back right directional language by composing sub-instructions. (``Turn around and walk forward'' for backward, ``Turn around and go to your right'' for back left, and ``Turn around and go to your left'' for back right). \figref{fig:directionintervention_clip} suggests for all three conditions, the agent may not be able to execute ``turn around'' or may not be able to compose ``turn around'' and other directional instructions. Similarly, estimating intervention effects as increased accumulated probability for backward ($3E{-}4$, $p=0.42$), back left ($-3E{-}3$, $p=0.33$), back right ($4E{-}3$, $p=0.38$). We observed overall similar effects for \textsc{Envdrop--imagenet} in \figref{fig:directionintervention_imagenet}, the agent responds to forward, left, and right strongly, but has no respond to backward, back left and back right. The estimated intervention effects are: forward (0.08, $p\approx0$), left (0.36, $p\approx0$), right (0.32, $p\approx0$), backward ($6E{-}4$, $p=0.27$), back left ($-5E{-}3$, $p=0.03$), back right ($-3E{-}3$, $p=0.29$)

\xhdr{Summary.} We find both \textsc{Envdrop--clip} and \textsc{Envdrop--imagenet} %
agents strongly respond to directional language for forward, left and right. But they are not able to respond to backward, back left, back right conditions properly. They only ground to part of the intervention instruction but fail on the whole instruction. (e.g., probability mass distributed to ``right'' for ``turn around and go to right'' instead of the correct direction, ``back left'') This may due to inability to parse ``turn around'' instructions. Some dataset biases from training are still evident in a bias towards forward actions.   

\subsection{Object}
\begin{figure}[t]
    \centering
    \includegraphics[width=1\linewidth]{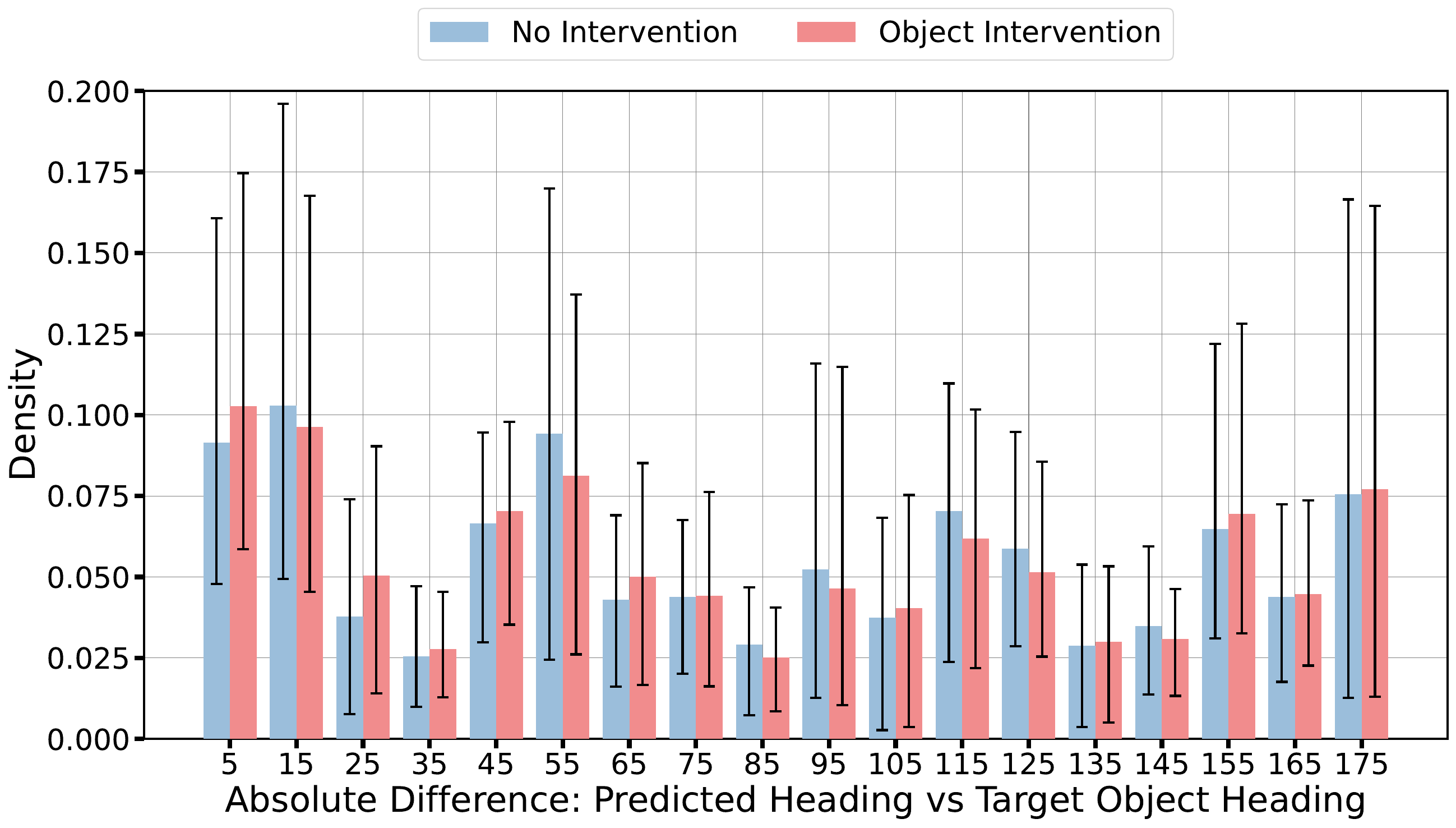}
    \caption{The distribution of the absolute difference between model prediction and target object direction for intervention and no intervention settings. (\textsc{Envdrop--clip})}
    \label{fig:obj_clip}
\end{figure}
\begin{figure}[t]
    \centering
    \includegraphics[width=1\linewidth]{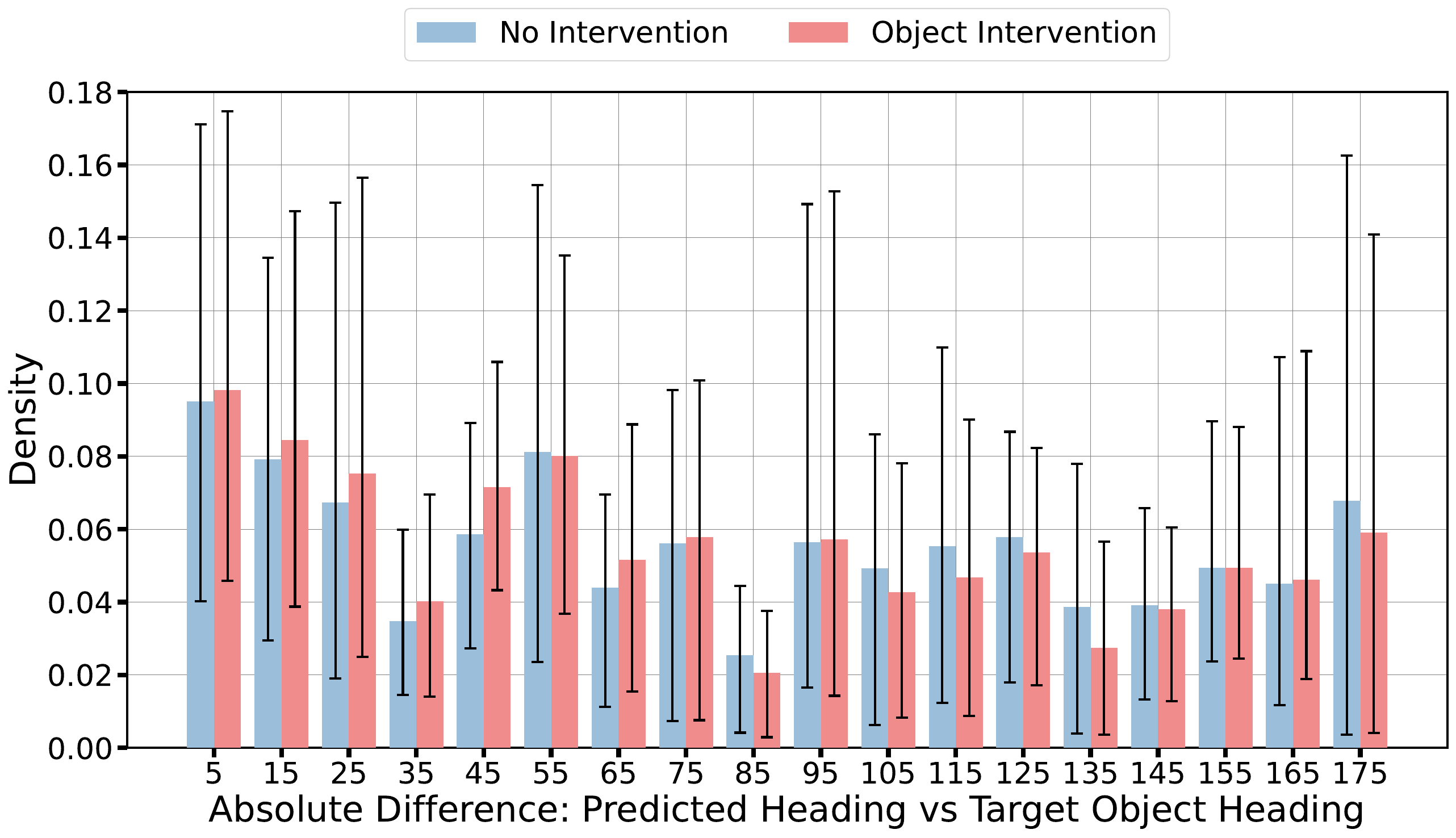}
    \caption{The distribution of the absolute difference between model prediction and target object direction for intervention and no intervention settings. (\textsc{Envdrop--imagenet})}
    \label{fig:obj_imagenet}
\end{figure}

\begin{figure}[t]
    \centering
    \includegraphics[width=1\linewidth]{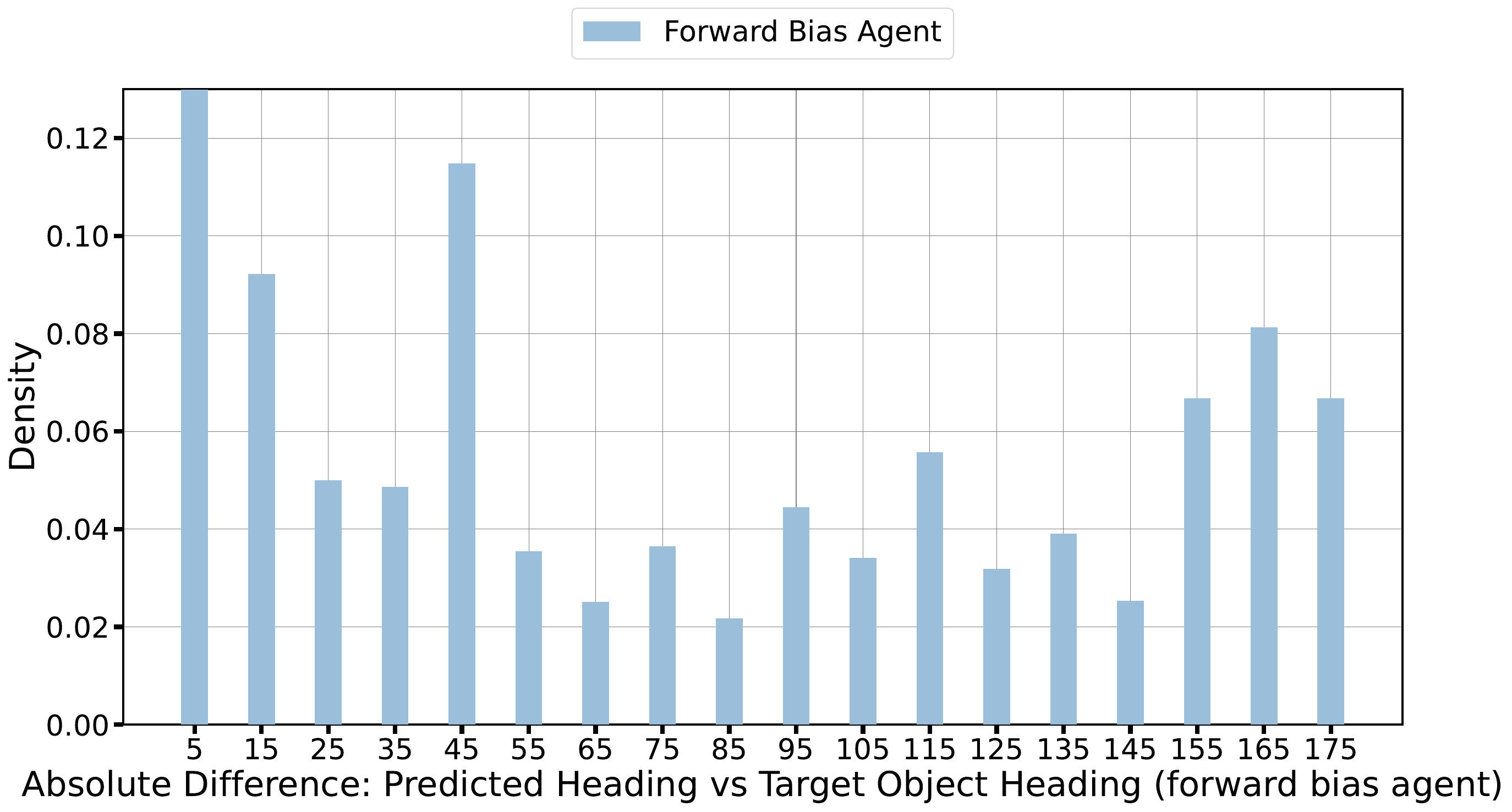}
    \caption{The distribution of absolute difference between model prediction and target object direction for forward bias agent.}
    \label{fig:obj_forward}
\end{figure}

\figref{fig:obj_clip} and \figref{fig:obj_imagenet} present distributions over angular distance for the intervention and no-intervention settings for \textsc{Envdrop--clip} and \textsc{Envdrop--imagenet} respectively.

For \textsc{Envdrop--clip}, the agent is not significantly responsive to object-seeking instruction. (blue \vs red bars in \figref{fig:obj_clip}. We again use a \texttt{lmer} to evaluate the effect of intervention
on the accumulated probability within 15 degree of absolute
angular difference. We find weak fixed
effect of $4E{-}2$ (\texttt{anova}, $p=2E{-}6$) for intervention vs non-intervention.
For \textsc{Envdrop--imagenet}(\figref{fig:obj_imagenet}), we find a weak fixed effect of $3E{-}2$, ($p=6E{-}7$) for intervention vs non-intervention. However, both \textsc{Envdrop--clip} and \textsc{Envdrop--imagenet} show a wide spread angular error that suggests the target objects are not being grounded accurately. (Recall all trajectories have neighboring nodes that would incur no more than 15 degrees of error.) To explore this error distribution further, we also examine a baseline \texttt{Forward bias} (\figref{fig:obj_forward} agent that places probability on neighbors inversely proportional to their relative heading. We find this baseline exhibits a similarly shaped error distribution to the agent -- suggesting the agent may be taking forward actions when uncertain about the target object.
As in our other experiments, the no intervention setting is more likely to stop than the intervention (50\% \vs 30\% for \textsc{Envdrop--clip}, 54\% \vs 37\% for \textsc{Envdrop-imagenet}.

\xhdr{Summary.} We find evidence for only a weak tendency to move towards referenced objects for \textsc{Envdrop--imagenet}, and \textsc{Envdrop--clip}.
\subsection{Room-seeking Instructions}

\begin{figure}[t]
    \centering
    \includegraphics[width=1\linewidth]{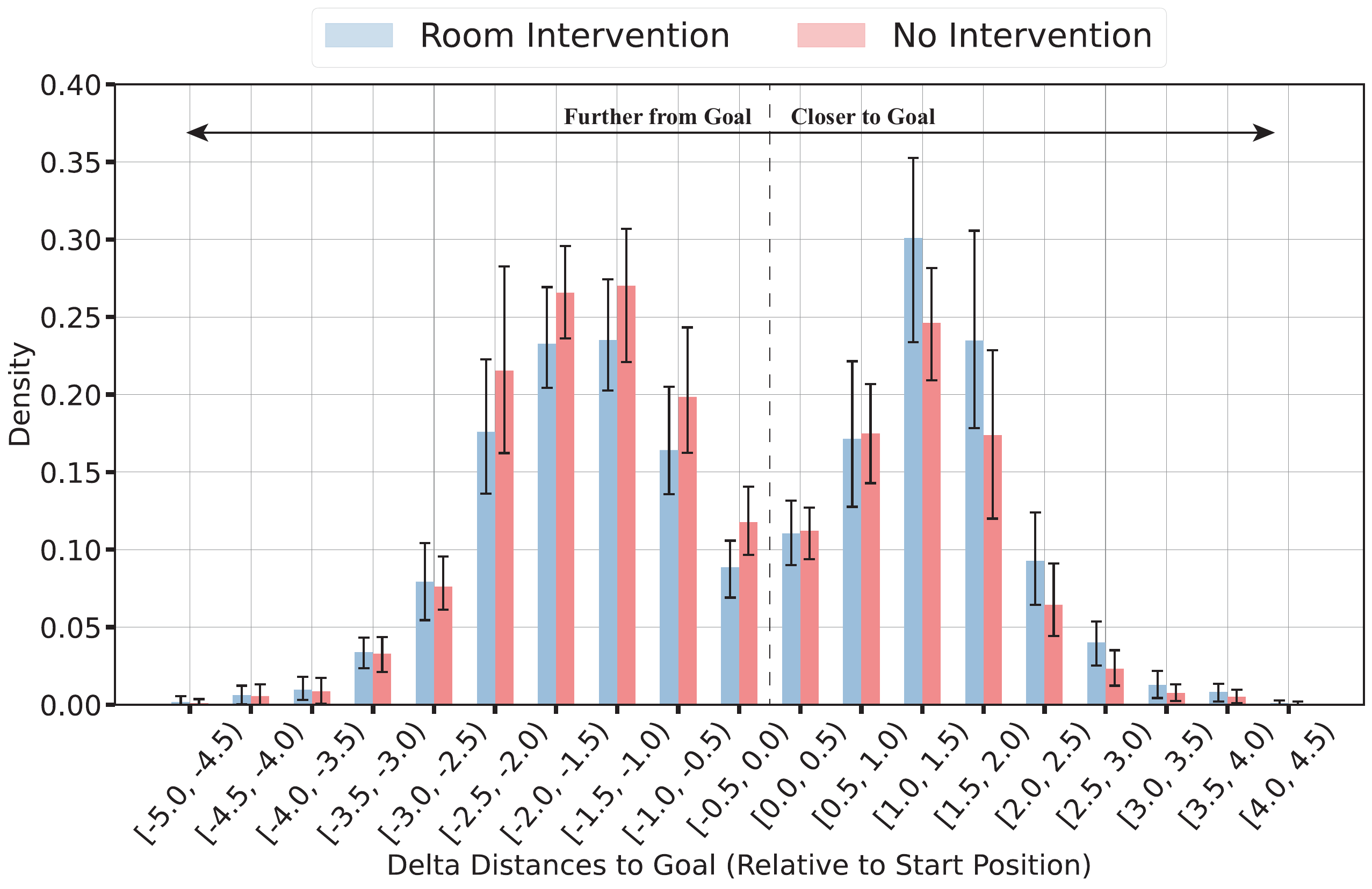}
    \caption{(\textsc{Envdrop--clip}) Distribution of delta distance to nodes of the target room type. The delta distance difference of distance to nodes of target room type (relative to start position) with or without intervention. Positive delta distance means the agent move closer to nodes of target type with intervention than otherwise. The distribution shift towards right with intervention than otherwise, indicates the agent is responsive to room-seeking instruction. (-0.15 \vs -0.41, $p=9E{-}5$ )}
    \label{fig:1hop_clip}
\end{figure}

\begin{figure}[t]
    \centering
    \includegraphics[width=1\linewidth]{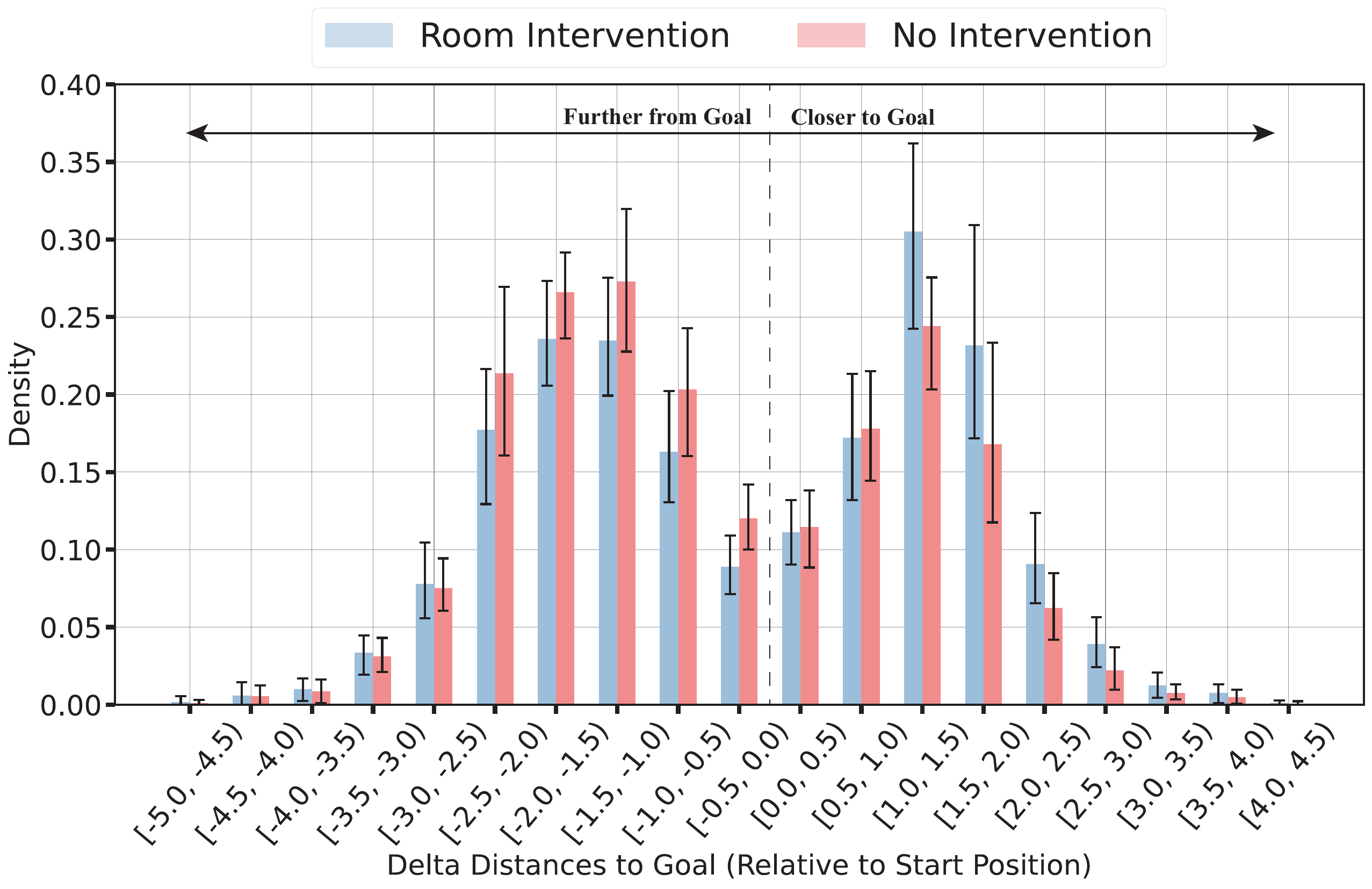}
    \caption{(\textsc{Envdrop--imagenet}) Distribution of delta distance to nodes of target room type. The delta distance difference of distance to the nodes of target room type (relative to start position) with or without intervention. Positive delta distance means the agent move closer to nodes of the target type with intervention than otherwise. The distribution shift towards right with intervention than otherwise, indicates the agent is responsive to room-seeking instruction. (-0.16 \vs -0.42, $p=4E{-}3$ )}
    \label{fig:1hop_imagenet}
\end{figure}

\xhdr{1-Hop Results.} The probabilities of delta distance for \textsc{Envdrop--clip} and \textsc{Envdrop--imagenet} are displayed in \figref{fig:1hop_clip} and \figref{fig:1hop_imagenet} respectively -- values greater than zero represent the agent moving \emph{closer} to nodes with the target room type. We observe a weak right-shift in the density suggesting the agents respond somewhat to the intervention. We again use a \texttt{lmer} to evaluate the effect of intervention on
the delta geodesic distance. For \textsc{Envdrop--clip}, we find the estimated fixed effect as 0.10 (\texttt{anova}, $p=9E{-}5$) for intervention vs no intervention. For \textsc{Envdrop-imagenet}, we find the estimated fixed effect as 0.05 ($p=4E{-}3$). However, Both the agents do not reliably place strong beliefs on neighbors with the target room type -- negative median delta distance and significant mass to the left of zero.

\xhdr{k-Hop Results.} We report the distance to the nearest node with target room type here.
\begin{figure}[t]
    \centering
    \includegraphics[width=1\linewidth]{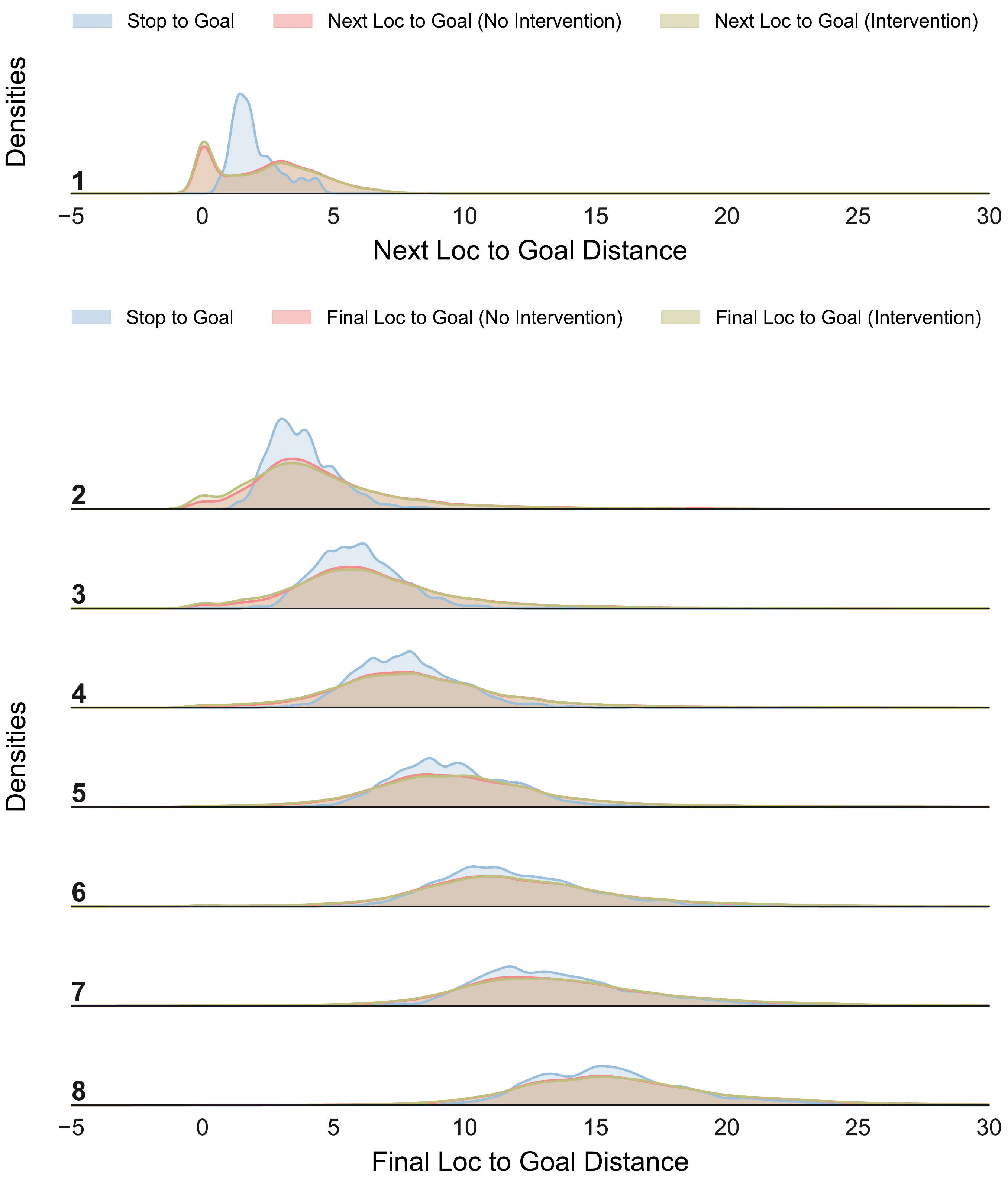}
    \caption{(\textsc{Envdrop--clip}) Distribution of geodesic distance to the nearest node of target room type for k-hop room-seeking experiments. \texttt{Stop to Goal} is a baseline agent that always takes the stop action.}
    \label{fig:ridgeplot_clip}
\end{figure}

\begin{figure}[t]
    \centering
    \includegraphics[width=1\linewidth]{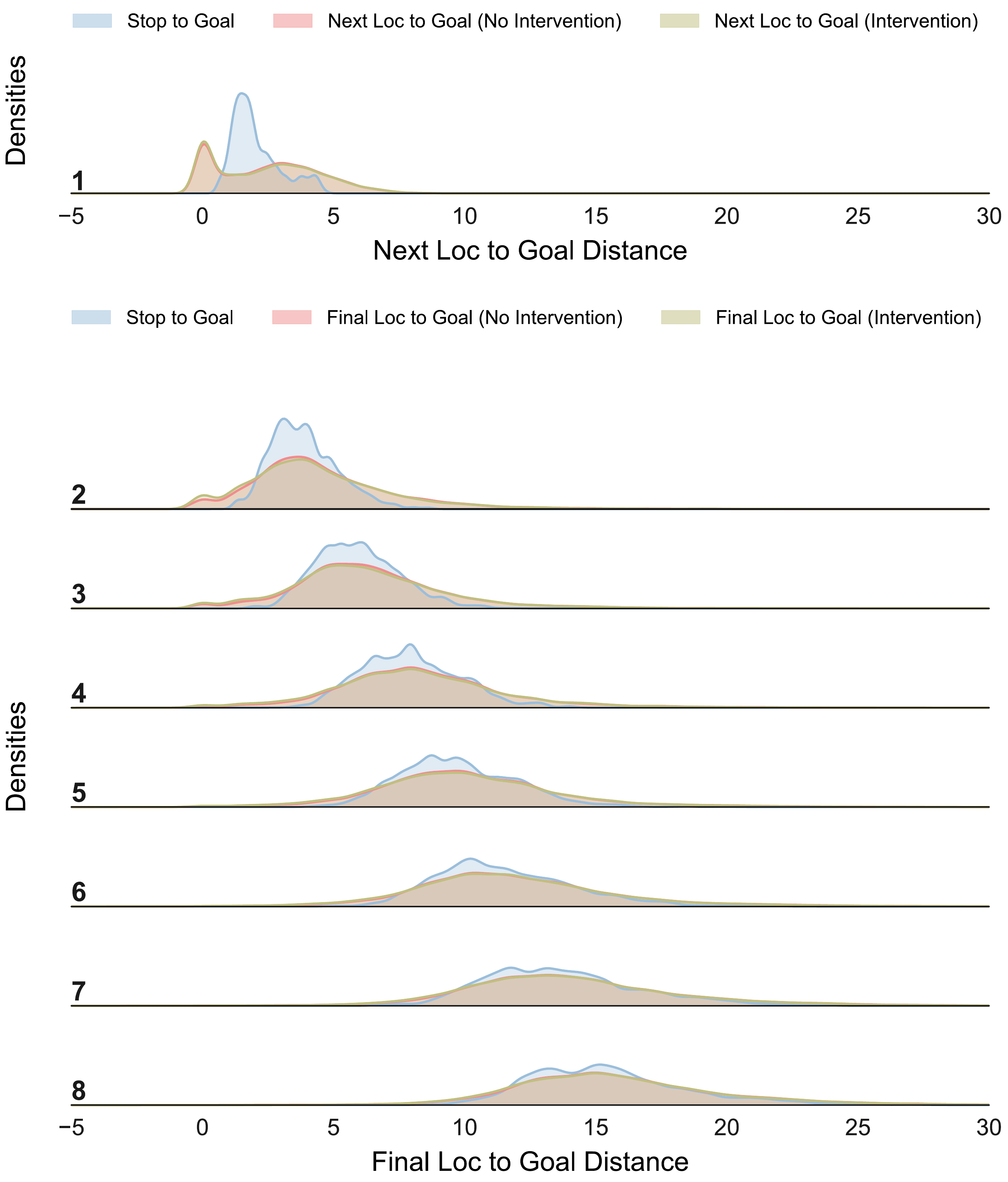}
    \caption{(\textsc{Envdrop--imagenet}) Distribution of geodesic distance to nearest node of target room type for k-hop room-seeking experiments. \texttt{Stop to Goal} is a baseline agent that always takes the stop action.}
    \label{fig:ridgeplot_imagenet}
\end{figure}

We show ridgeline plots in \figref{fig:ridgeplot_clip} and \figref{fig:ridgeplot_imagenet} for \textsc{Envdrop--clip} and \textsc{Envdrop--imagenet}, respectively. We compare distance to the nearest node of target room type distributions for 1- to 8-hops. %
For both agents, we find the error increases with target room distance. We again leverage a \texttt{lmer} to evaluate the effect of intervention on $d_{geo}(n_{end}, n_{near})$. For \textsc{Envdrop--imagenet}, we find weak effect of $\leq-0.08$ (\texttt{anova}, $p=8E{-}3$) for intervention \vs non-intervention for 2--8 hops with 95\% confidence). For \textsc{Envdrop--clip}, we find similar weak effects $\leq-0.08$ ($p=1E{-}2$) for 2--6 hops with 95\% confidence. Overall, this suggests agents have limited ability to search for rooms based on common sense exploration.

\xhdr{Summary.} Both the \textsc{Envdrop--clip}~\cite{shen2021much} and \textsc{Envdrop--imagent}~\cite{tan2019learning}
agents are only weakly sensitive to room type reference instructions when the room is visible (within one hop) but lack the ability to perform common sense exploration to find further away rooms (k-hop). Overall sensitivity is low, suggesting the agent may not rely on room-specifying portions of instructions when navigating.

%% file: supplementary/templates.tex
\section{Templates and Examples}
\tabref{tab:templates} show templates we used in our cases studies. And as we mannully designed the templates from examining RxR \cite{ku2020room} dataset, we also provide example instructions in \tabref{tab:example_of_templates} containing the templates. Note the we make sure each part of templates can be found in training data.%
\clearpage
\onecolumn
\begin{table}[]
\centering
\resizebox{\textwidth}{!}{%
\begin{tabular}{p{0.4\textwidth}p{0.5\textwidth}}
\hline
\textbf{Skill-specific language} &
  \textbf{Template} \\ \hline
Stop Instruction &
  \begin{tabular}[c]{@{}l@{}}This is your destination.\\ This is your end point.\\ You reached your destination.\\ You are done.\end{tabular} \\ \hline
Unconditional Directional Instructions &
  \begin{tabular}[c]{@{}l@{}}Walk forward. (\textit{forward})\\ Turn around and walk forward. (\textit{backward})\\ Turn left and walk forward. (\textit{left})\\ Turn right and walk forward. (\textit{right})\\ Turn around and go to your right. (\textit{back left})\\ Turn around and go to your left. (\textit{back right})\end{tabular} \\ \hline
Object-seeking Instruction &
  Walk towards the XX (\textit{Object})\\ \hline
Room-seeking Instruction &
  Walk towards the XX (\textit{room type}) \\ \hline
\end{tabular}%
}
\caption{Templates for Skill-specific language used for our work.}
\label{tab:templates}
\end{table}

\begin{longtable}[c]{p{0.35\textwidth}p{0.6\textwidth}}
\hline
\textbf{Template} &
  \textbf{Examples} \\ \hline
\endfirsthead
\endhead
\begin{tabular}[c]{@{}l@{}}
This is your destination.\\ This is your end point.\\ You reached your destination.\\ You are done. 
\end{tabular}
& 
As you're facing the wall, you're gonna see 4 white coats to your right. Just turn around and take a few steps forward, you're gonna have a small sink to your left. \textbf{This is your destination.}
  \\ \hline
Walk forward. &
  Starting off in a side room library. \textbf{Walk forward.} And you ll see an open living room with chairs, a piano to your left, a desk to your right , wall is also to the right. Continue straight to the middle of the chair and the desk. Once the desk is to your right forward is a window with a mountain range and in front is also another couch, a big long couch. And to the left is a small circle table. Taking one last step and onto the couch the table is still to your left \\ \hline
Turn around &
  You will start by standing in front of a glass door and on your right is a doorway. \textbf{Turn around} and you will see a doorway to the washroom. Walk towards the doorway and inside the washroom. Once you're there, stand in between the sink and the bathtub and once you're there, you're done. \\ \hline
Go to your right &
  You are facing a large window.  You are going to turn all the way around.  You are going to exit this room and make a right.  You are going to move forward into this room on the large blue rug.  And you are going to go to the middle door on your right.  The doors will now be open and you are going to take a step outside.  You are going to \textbf{go to your right.}  You are going to move forward down this pathway.  And you are going to stop when you are right next to the yellow outlined glass window on the building will be on your right and on your left is just going to be the cement banister between 2 columns and you are done. \\ \hline
Go to your left &
  You are facing an open door and a massage bed.  You are going to go thru the door. And once thru the door you are going to make an immediate right.  You are going to step into this room and you are going to \textbf{go to your left.}  You are going to hop over to the 3rd massage lounger on your right.  Then you are going to make a right and go thru the entrance.  You are going to continue moving forward, you will see a staircase in front of you.  And you are going to stop right when you are near the banister to the staircase, on the left of you is going to be a corner with a statue and to the right of you is going to be a seating area with 2 wicker chairs and you are done. \\ \hline
Turn left &
  Begin facing some shelves. Turn around and head out the open doors. Head to the dining table and \textbf{turn left.} Head down the left side of the dining table until you reach the living area. \textbf{Turn left} and go to the random swing from here head to the white chair in the corner of the room on the elevated platform under the odd art and you're done. \\ \hline
Turn right &
  You're in a living room. \textbf{Turn right} and you'll see a small hallway. Go into it. Toward the doors you can see that are horizontally slatted with wood. In the hallway on the left you will see a table with lots of photos on it. Go toward the table. Look at the table look right. You'll see another room in the distance with a large rectangular table with various boxes and a lamp on it. Step toward there, you'll see its a bedroom. Step to the foot of the bed, look right walk over to the single chair to the right of the bed. Step into the corner left of that chair and stop. \\ \hline
Walk towards the XX (\textit{object}) &
  We are standing inside an empty walk in closet. We are going to head out inside the bedroom. \textbf{Walk towards the bed} and outside on the balcony. Stop when you're outside on the balcony overlooking the city. That's it. \\ \hline
Walk towards the XX (\textit{room type}) &
  You're in a bathroom with wooden floors and wooden walls. There's a bathtub in front of you. Walk around the bathtub. To your right you see a toilet, a cabinet, a sink and a mirror. In front of you there's a doorway exiting the bathroom. Walk towards this doorway. Continue to walk towards the door. Exit outside into the main room. You're now in the main room. It also has wooden walls and wooden floors. There's a kitchen in front of you. \textbf{Walk towards the kitchen.} You're now in the kitchen, to your right you can see some cabinets, a sink, a table and you've reached the end. \\ \hline
\caption{Examples of templates from RxR \cite{ku2020room} training dataset.}
\label{tab:example_of_templates}\\
\end{longtable}
\clearpage
\twocolumn

%% file: supplementary/teacher_forcing_effects.tex
\begin{table}[t]
\centering
\resizebox{\columnwidth}{!}{%
\begin{tabular}{lcccccc}
\hline
Method  & NE   & OE   & SR    & SPL   & nDTW  & sDTW  \\ \hline
HAMT    & 7.75 & 5.48 & 42.49 & 39.33 & 54.01 & 35.05 \\
HAMT--tf & 4.92 & 3.43 & 52.94 & 50.75 & 72.41 & 47.98 \\ \hline
\end{tabular}%
}
\caption{We report scores for teacher forcing part of ground truth (HAMT--tf) vs No teacher force (HAMT). We find by forcing the agent until the end of partial ground truth, there is no performance drop but increase across all metrics than otherwise.}
\label{tab:tf_effect}
\end{table}

\section{Teacher Forcing Effects}
We run a small experiment to verify agents continue to behave rationally after being forced through the intervention trajectories. Consider a truncated $(\tau, I)$, pair, we replace the $I$ with full instruction $I_f$. Given full instruction $I_f$, agents were either forced until the final node of $\tau$ then started to take argmax actions, or without teacher forcing along $\tau$ at all. \tabref{tab:tf_effect} indicates no performance drop occurred due to the teacher forcing process. (Perhaps unsurprisingly, teacher forcing brings a 10\% performance increase.)